%% file: main.tex
\documentclass{article}

\usepackage[nonatbib, final]{neurips_2022}

\usepackage{wrapfig}

\usepackage[utf8]{inputenc} 
\usepackage[T1]{fontenc}    
\usepackage{hyperref}       
\usepackage{url}            
\usepackage[table]{xcolor}
\usepackage{subcaption}
\usepackage{booktabs}       
\usepackage{multirow}
\usepackage{amsfonts}       
\usepackage{nicefrac}       
\usepackage{microtype}      
\usepackage{xcolor}         
\usepackage{amsmath}
\usepackage{graphicx}
\usepackage{enumitem}
\usepackage{longtable}
\usepackage{bm}
\usepackage{ulem}
\usepackage{mathtools}

\newcommand{\pname}[1]{\textcolor{black}{SELOR}}

\newcommand{\reason}[1]{\textbf{\textcolor{brown}{#1}}}
\definecolor{aftercolor}{HTML}{548235}
\newcommand{\after}[1]{\textbf{\textcolor{aftercolor}{#1}}}

\title{
Self-explaining deep models with logic rule reasoning
}

\author{
    Seungeon Lee\thanks{
    Work done during internship at Microsoft Research Asia} \\
    KAIST School of Computing \\
    IBS Data Science Group \\
    \texttt{archon159@kaist.ac.kr} \\
    \And
    Xiting Wang\thanks{Corresponding Author} \\
    Social Computing Group \\
    Microsoft Research Asia \\
    \texttt{xitwan@microsoft.com} \\
    \And
    Sungwon Han\footnotemark[1] \\
    KAIST School of Computing\\
    IBS Data Science Group \\
    \texttt{lion4151@kaist.ac.kr} \\
    \And
    Xiaoyuan Yi \\
    Social Computing Group \\
    Microsoft Research Asia \\
    \texttt{xiaoyuanyi@microsoft.com} \\
    \And
    Xing Xie \\
    Social Computing Group \\
    Microsoft Research Asia \\
    \texttt{xing.xie@microsoft.com} \\
    \And
    Meeyoung Cha\footnotemark[2] \\
    IBS Data Science Group \\
    KAIST School of Computing\\
    \texttt{mcha@ibs.re.kr} \\
}

\begin{document}

\normalem
\maketitle

\input{source/0abstract}
\input{source/1introduction}

\input{source/4model_design}
\input{source/5experiments}
\input{source/6conclusion}

\input{source/acknowledgement}

\bibliographystyle{unsrt}
\bibliography{reference}

\newpage
\input{source/checklist}
\newpage
\input{source/appendix}

\end{document}

%% file: source/0abstract.tex
\begin{abstract}

We present \pname{}, a framework for integrating self-explaining capabilities into a given deep model to achieve both high prediction performance and human precision.
By ``human precision'', we refer to the degree to which humans agree with the reasons models provide for their predictions. Human precision affects user trust and allows users to collaborate closely with the model.
We demonstrate that logic rule explanations naturally satisfy human precision with the expressive power required for good predictive performance.
We then illustrate how to enable a deep model to predict and explain with logic rules.
Our method does not require predefined logic rule sets or human annotations and can be learned efficiently and easily with widely-used deep learning modules in a differentiable way.
Extensive experiments show that our method gives explanations closer to human decision logic than other methods while maintaining the performance of deep learning models.

\end{abstract}

%% file: source/1introduction.tex
\section{Introduction}
\label{sec:introduction}
Deep learning has shown high predictive accuracy in a wide range of tasks, but its inner working mechanisms are obscured by complex model designs. 
This raises important questions about whether a deep model is ethical, trustworthy, or capable of performing as intended under various conditions~\cite{ribeiro2016should}.\looseness=-1

Many approaches have been proposed to help humans assess and comprehend model decisions.
Recent work on explainability has primarily focused on providing \textbf{post-hoc} explanations for black-box models that have already been trained~\cite{thrun1994extracting,ribeiro2016nothing,lei2016rationalizing,alikaniotis2016automatic,strobelt2017lstmvis,murdoch2018beyond,peake2018explanation,liang2020adversarial,gao2021learning}.
Post-hoc methods do not change the model and hence preserve the predictive performance while providing the additional benefit of explainability.
These methods have achieved considerable success in providing valuable insights for model understanding, 
but there are also known challenges such as computational cost~\cite{alvarez2018towards} and trust issues~\cite{rudin2019stop}.
For example, many popular post-hoc methods test the complex black-box model thousands of times to obtain a complete and faithful understanding of the model around a single instance~\cite{ribeiro2016should,lundberg2017unified,ribeiro2018anchors}.
Subroutines such as full optimization or reverse propagation are generally required, 
introducing approximations or heuristic assumptions that may lead to misinterpretation~\cite{ribeiro2018anchors,guan2019towards}.
Because there is no guarantee that explanations are always faithful to the model~\cite{rudin2019stop}, there exists a ``general uneasiness'' among practitioners about using and trusting post-hoc explanations~\cite{hong2020human}.
\textbf{Self-explaining} models naturally solve these issues, making them an ideal choice when interpretability can be considered from the model design phase~\cite{letham2015interpretable,yang2017scalable,evans2018learning,angelino2017learning}.
Because the explanation mechanism is integrated inherently, these models can predict and explain simultaneously with a single forward propagation without any approximations or heuristic assumptions that decrease the faithfulness of explanations. 
Self-explaining methods may also improve robustness~\cite{alvarez2018towards} and provide actionable insights for directly refining model parameters without having to calibrate the dataset~\cite{ming2019interpretable,chen2020towards}.\looseness=-1

Based on these observations, we regard self-explaining models as providing a stronger link between humans and machine learning models, reducing misunderstanding and allowing direct control of the model based on human insights.
The main challenge in achieving this new level of 
human-machine
collaboration then becomes how to ensure self-explaining models' precision both in terms of \textbf{predictive performance} and \textbf{human precision}.
Human precision refers to whether models' explanations of decision-making processes align with human decision logic.
Existing approaches ensure explanations to be \emph{easy to read}, for example, by requiring explanations to be simple and smooth in a local area~\cite{alvarez2018towards}.
However, there is little guarantee that a given explanation is a \emph{correct} rationale for prediction according to human perception.
For example, the explanation ``\emph{awesome}$\geq$2'' (i.e., the word ``\emph{awesome}'' appears twice in reviews) is a good rationale for positive sentiment, while ``\emph{is}$\geq$1'' $\Rightarrow$ \emph{positive sentiment} is easy to read but unreasonable to humans.
Without insurance for human precision, users may constantly find unreasonable explanations, which can significantly hamper user trust and prevent them from identifying actionable insights 
for model refinement.
An interesting research question, then, is: how can self-explaining models generate explanations that are consistent with human decision logic?\looseness=-1

\begin{figure}[t]
    \vspace{-2mm}
    \centering
    \begin{subfigure}[t]{0.38\textwidth}
        \centering
        \includegraphics[width=\textwidth]{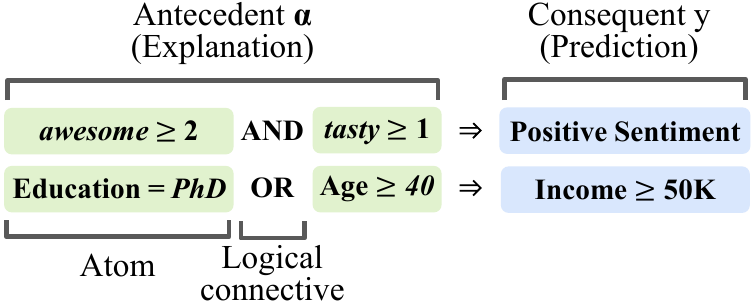}
        \caption{Rule example}
        \label{fig:rule_example}
    \end{subfigure}
    \hspace{1mm}
    \begin{subfigure}[t]{0.20\textwidth}
        \centering
        \includegraphics[width=\textwidth]{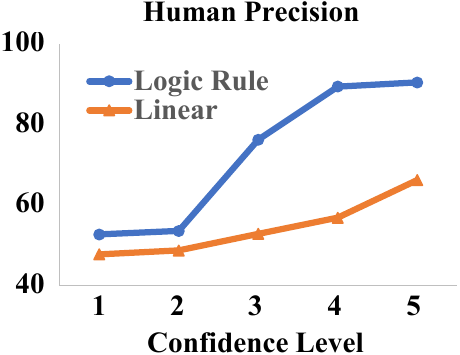}
        \caption{Human precision}
        \label{fig:rule_human_precision}
    \end{subfigure}
    \hspace{1mm}
    \begin{subfigure}[t]{0.38\textwidth}
        \centering
        \includegraphics[width=\textwidth]{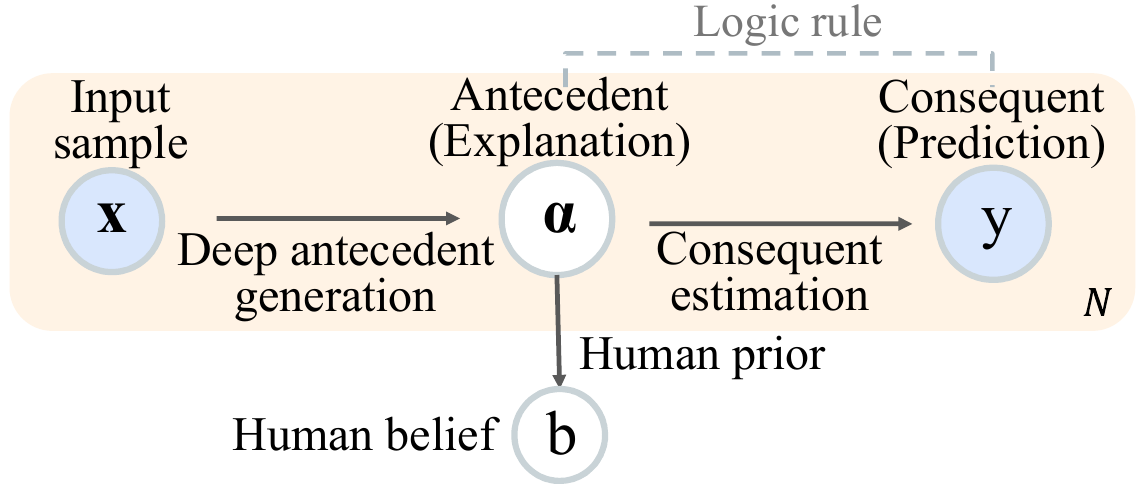}
        \caption{Our generative process}
        \label{fig:rule_logic}
    \end{subfigure}
    
    \caption{Reasoning with logic rules: (a) examples of logic rule explanations; (b) human precision for logic rule and linear regression explanations~\cite{alvarez2018towards};  (c) generative process of our logic rule reasoning.\looseness=-1}
    \vspace{-6mm}
    \label{fig:rule_intro}
\end{figure}

To answer this question, we need to decide what information models obtain from humans.
Collecting ground-truth labels of human decision processes for every input instance~\cite{kulesza2015principles, schramowski2020making, lertvittayakumjorn2020find, ciravegna2021human, stammer2021right, bontempelli2021toward} is expensive and limits the method's scalability.
Moreover, forcing the model to make decisions exactly like humans may be unwise since it could limit its data learning capability or even learn human biases that may significantly decrease the model's performance.
To address this issue, it is important that humans provide guidance at a higher level that allows the models to learn freely based on data.
Accordingly, we propose two desirable properties for human precision.
The first property, \textbf{global coherency}, restricts the explanation form to be consistent with human reasoning logic, thereby minimizing the probability of misinterpretation.
The second property, \textbf{local coherency}, requires that each explanation naturally lead to the prediction according to human perception, thereby making explanations a correct rationale for the model output.
As humans can hardly provide guidance for each explanation, it is more desirable that the models can automatically guarantee local coherency based on human guidance on global coherency.
\looseness=-1

A key to satisfying these two properties is logic rules.
As shown in Fig.~\ref{fig:rule_example}, logic rules can have flexible forms that meet human logic and preferences, making them easy to satisfy global coherency.
For example, the logical connectives can be traditional (e.g., AND, OR, and NOT) or self-defined (e.g., BEFORE).
Moreover, the logic rules explicitly model whether an explanation can lead to a prediction by testing the hypothesis across the entire dataset. 
This ensures a meaningful relationship between explanations and predictions that leads to local coherency. 
Fig.~\ref{fig:rule_human_precision} shows that logic rule explanations achieve even higher human precision than linear-regression-based explanations with local stability, while providing a confidence score that correlates with human precision (more details in Appendix~\ref{sec:intro_sup}).
Lastly, logic rules of different logical connectives correspond to a diverse set of feature interactions, providing the expressive power for good predictive performance.
\looseness=-1

This paper proposes \textbf{\pname{}}, a framework for upgrading a deep model with a \textbf{S}elf-\textbf{E}xplainable version with \textbf{LO}gic rule \textbf{R}easoning 
capability.
Our work is inspired by neuro-symbolic reasoning~\cite{de2020statistical}, which integrates deep learning with logic rule reasoning to inherit advantages from both. The most related works in this discipline are neural-guided search that finds a global logic program that works for (most) input-output pairs~\cite{evans2018learning, valkov2018houdini, ellis2018learninglibraries, kalyan2018neural}, or identifies a local logic program and rule for the given instance~\cite{ellis2018learningto, mao2019neuro, yu2022probabilistic, wang2022multi, yang2017differentiable}. We adopt the latter paradigm, as global explanations for deep models usually fails to possess the same predictive power that is comparable with the deep models~\cite{ribeiro2018anchors}. Existing works for generating local programs or rules have achieved promising results by effectively leveraging instance-level guidance about local programs or rules~\cite{ellis2018learningto, mao2019neuro, yu2022probabilistic}, strong external knowledge such as knowledge graphs~\cite{wang2022multi, yang2017differentiable, zhao2020leveraging}, and a small set of predefined rules~\cite{okajima2019deep}. However, in our scenario, there is no instance-level guidance about the ground-truth rules, and leveraging strong knowledge such as a small set of predefined rules may introduce bias into the deep networks, as shown in our experiment results of RCN~\cite{okajima2019deep}. To address this, we propose a logic rule reasoning framework that leverages global level human priors about rules (e.g., desirable form and property of candidate atoms) and generate explanations by optimizing rule confidence, which can be automatically computed based on the training data. Moreover, we design a neural consequent estimator that can accurately approximate the confidence even for rare rules and combine it with recursive Gumbel-Softmax~\cite{jang2017categorical} to search the solution space effectively. Codes are released at Github.\footnote{https://github.com/archon159/SELOR}
\looseness=-1

Our main contributions are as follows.
\begin{itemize}[nosep,leftmargin=1em,labelwidth=*,align=left]
    \item 
    Our work suggests that human precision is key for self-explaining models to bridge human logic and model decision logic seamlessly. 
    Logic rule-based explanations enable high human precision while allowing the expressive power to achieve high prediction performance.\looseness=-1
   \item 
   We propose a logic rule reasoning framework that upgrades a given deep model into a self-explainable version by naturally integrating human priors, rule confidence modeling, and rule generation as an essential part of model prediction. Our method can achieve high human precision without depending on strong external knowledge, such as instance-level guidance about rules, knowledge graphs, or a small number of rule candidates.\looseness=-1
   \item 
    Numerical experiments and user studies confirm key strengths of our framework in terms of human precision and robustness against noisy labels with maintenance of prediction performance.\looseness=-1
\end{itemize}

%% file: source/4model_design.tex
\section{Deep Logic Rule Reasoning}
\label{sec:design}

\subsection{Formulation of Logic Rules}
\label{sec:logic_formulation}
A logic rule $\bm{\alpha} \Rightarrow y$, as shown in Fig.~\ref{fig:rule_example}, consists of an antecedent $\bm{\alpha}$ and a consequent $y$.
Meanings of symbols used in this paper are defined in Appendix~\ref{sec:symbol}.

\begin{itemize}[nosep,leftmargin=1em,labelwidth=*,align=left]

\item An \textbf{antecedent} $\bm{\alpha}$ is the condition to apply the rule and corresponds to an explanation in a logic form. It is represented as a sequence $\bm{\alpha}=(o_1...,o_L)$, where $o_i$ is 
either an atom or a logical connective.\looseness=-1

\begin{itemize}[nosep,leftmargin=1em,labelwidth=*,align=left]

    \item An \textbf{atom} is the smallest unit of explanation that corresponds to a single \emph{interpretable feature} of a given input (e.g., ``\emph{awesome}$\geq$2''). The interpretable features may be different from those in deep learning models. They could, for example, have a different granularity (e.g., words or phrases) than the model features (e.g., partial words), be a statistical feature (e.g., word frequency), or be derived using external tools (e.g., grammatical tagging of a word). Mathematically, each atom $o_i$ is a Boolean-value function, with $o_i(\mathbf{x})$ returning true if the $i$-th interpretable feature is present in input $\mathbf{x}$ and false, otherwise. More detail about atom selection is in Appendix~\ref{sec:implementation_details}\looseness=-1

    \item A \textbf{logical connective} combines atoms to form an explanation. Logical connectives can be traditional ones like AND, OR, NOT, or self-defined ones, as long as they take one or more Boolean values as the input and output a single Boolean value.\looseness=-1
    
\end{itemize}

We say that an input sample $\mathbf{x}$ \textbf{satisfies} an antecedent $\bm{\alpha}$, if $\bm{\alpha}(\mathbf{x})$ is true. 

\item The \textbf{consequent} $y$ is the model's prediction output given the antecedent.
For example, $y$ is the predicted class in a classification task, whereas $y$ is an explicit number in a regression task. We mainly consider classification in the paper and extend the cases to regression in Appendix~\ref{sec:regression}.
\looseness=-1
\end{itemize}

\subsection{Framework for Deep Logic Rule Reasoning}
\label{sec:framework_logic}
 
Let us denote $f$ as a deep learning model that estimates probability $p(y|\mathbf{x})$, where $\mathbf{x}$ is the input data sample and $y$ is a candidate class.
We upgrade model $f$ to a self-explaining version by adding a latent variable $\bm{\alpha}$, 
which is an explanation in the logic form. 
Then, we can reformulate $p(y|\mathbf{x})$ as \looseness=-1
\begin{equation}
    \label{eq:prob_decomposition_1}
    p(y|\mathbf{x},b) = 
    \sum_{\bm{\alpha}}  
    p(y|\bm{\alpha},\mathbf{x},b)p(\bm{\alpha}|\mathbf{x},b)
    = 
    \sum_{\bm{\alpha}}  
    p(y|\bm{\alpha})p(\bm{\alpha}|\mathbf{x},b),
\ \ \ s.t., \ \ \ 
    \Omega(\bm{\alpha}) \leq S
\end{equation}
Here, $b$ represents a human's prior belief about the rules, e.g., the desirable form of atoms and logical connectives, $\Omega(\bm{\alpha})$ is the required number of logic rules to explain given input $\mathbf{x}$, and $S$ is the number of samples (logic rules chosen by the model).
Eq. (\ref{eq:prob_decomposition_1}) includes two constraints essential for ensuring explainability.
The first constraint $p(y|\bm{\alpha},\mathbf{x},b)=p(y|\bm{\alpha})$ requires that explanation $\bm{\alpha}$ contains all information in the input $\mathbf{x}$ and $b$ that is useful to predict $y$.
Without the constraint, the model may ``cheat'' by predicting $y$ directly from the input instead of using the explanation (more details in Appendix~\ref{sec:probability_decomposition}).
The second constraint $\Omega(\bm{\alpha}) \leq S$ requires that the model can be well explained by using only $S$ explanations, where $S$ is small enough to ensure readability ($S=1$ in our implementation).
\looseness=-1

We can further decompose Eq.~(\ref{eq:prob_decomposition_1}) based on the independence between the input $\mathbf{x}$ and the human prior belief $b$, following the generative process in Fig.~\ref{fig:rule_logic} (proof and assumptions in Appendix~\ref{sec:probability_decomposition}):
\looseness=-1
\begin{equation}
    \label{eq:prob_decomposition_2}
    p(y|\mathbf{x},b) =
    \sum_{\bm{\alpha}}  
    p(y|\bm{\alpha})p(\bm{\alpha}|\mathbf{x},b)
    \propto
         \sum_{\bm{\alpha}} 
     \underbrace{p(b | \bm{\alpha})}_{\substack{\text{Human}\\ \text{prior}}} \cdot
\underbrace{p(y | \bm{\alpha})}_{\substack{\text{Consequent}\\ \text{estimation}}} \cdot\ \ 
    {\underbrace{p(\bm{\alpha} | \mathbf{x})}_{\mathclap{\substack{\text{\ Deep antecedent}\\ \text{\ generation}}}}}\ ,
\ \ \ s.t., \ \ \ 
    \Omega(\bm{\alpha}) \leq S
\end{equation}
The three derived terms correspond to three main modules of the proposed framework, \pname{}:
\begin{itemize}[nosep,leftmargin=1em,labelwidth=*,align=left]
    \item \textbf{Human prior $p(b | \bm{\alpha})$} 
    specifies human guidance regarding desirable forms for rules to minimize the probability of misunderstanding and ensure global coherency (Sec.~\ref{sec:prior}).

    \item \textbf{Consequent estimation} $p(y | \bm{\alpha})$  ensures a meaningful and consistent relationship between the explanation $\bm{\alpha}$ and prediction $y$, so that each explanation naturally leads to the prediction according to human perception and satisfies local coherency (Sec.~\ref{sec:consequent_esimation}).\looseness=-1

   \item \textbf{Deep antecedent generation $p(\bm{\alpha} | \mathbf{x})$} uses the deep representation of input $\mathbf{x}$ learned by the given deep model $f$ to find an explanation $\bm{\alpha}$ that maximizes global and local coherency (Sec.~\ref{sec:rule_generator}).
\end{itemize}

The sparsity constraint $\Omega(\bm{\alpha}) \leq S$ for the explanations can be enforced by sampling from $p(\bm{\alpha} | \mathbf{x})$.
In particular, we rewrite Eq.~(\ref{eq:prob_decomposition_2}) as an expectation and estimate it through sampling:
\begin{align}
    \label{eq:prob_approximation}
    p(y|\mathbf{x},b)
    \propto 
    \sum_{\bm{\alpha}}p(b | \bm{\alpha})\ p(y | \bm{\alpha})\  p(\bm{\alpha} | \mathbf{x})
    = 
    \mathop{{}\mathbb{E}}_{\mathclap{\bm{\alpha} \sim \atop p(\bm{\alpha} | \mathbf{x})}} \ p(b | \bm{\alpha})p(y | \bm{\alpha}) 
    \approx {1 \over S} \sum_{\mathclap{
    s\in[1,S] \atop {\bm{\alpha}^{(s)}\sim p(\bm{\alpha} | \mathbf{x})}}}  \ p(b | \bm{\alpha}^{(s)})\ p(y | \bm{\alpha}^{(s)})
\end{align}
where $\bm{\alpha}^{(s)}$ is the $s$-th sample of $\bm{\alpha}$. 
For example, to maximize the approximation term with $S=1$, the explanation generator $p(\bm{\alpha}|x)$ must find a single sample $\bm{\alpha}^{(s)}$ that yields the largest $p(b|\bm{\alpha}^{(s)})p(y|\bm{\alpha}^{(s)})$, and it
 needs to assign a high probability to the best $\bm{\alpha}^{(s)}$.
Otherwise, other samples with a lower $p(b|\bm{\alpha}^{(s)})p(y|\bm{\alpha}^{(s)})$ may be generated, thereby decreasing $p(y|\mathbf{x},b)$. 
This ensures the sparsity of $p(\bm{\alpha}|\mathbf{x})$ and the model interpretability. 
If there are multiple best explanations that result in the exact same $p(b|\bm{\alpha}^{(s)})p(y|\bm{\alpha}^{(s)})$, the explanation generator may find all of them.
\looseness=-1

\subsection{Human Prior \texorpdfstring{$p(b | \bm{\alpha})$}{Lg}}
\label{sec:prior}
Human prior $p(b | \bm{\alpha})=p_h(b | \bm{\alpha})p_s(b | \bm{\alpha})$ consists of hard priors $p_h(b | \bm{\alpha})$ and soft ones $p_s(b | \bm{\alpha})$.

\textbf{Hard priors} categorize the feasible solution space for the rules: $p_h(b | \bm{\alpha})=0$ if $\bm{\alpha}$ is not a feasible solution.
Humans can easily define hard priors by choosing the atom types, such as whether the interpretable features are words, phrases, or statistics like word frequency. The logical connectives to be considered (e.g., AND, NOT) can also be chosen, as well as the antecedent's maximum length $L$. \pname{} does not require a predefined rule set.
Nonetheless, we allow users to enter one if it is more desirable in some application scenarios. 
A large solution space increases the time cost for deep logic rule reasoning (Sec.~\ref{sec:optimization}) but also decreases the probability of introducing undesirable bias.
\looseness=-1

\textbf{Soft priors} model different levels of human preference for logic rules.
For example, people may prefer shorter rules or high-coverage rules that satisfy many input samples. 
The energy function can parameterize such soft priors:
$p_s(b | \bm{\alpha}) \propto \text{exp}(-\mathcal{L}_b(\bm{\alpha}))$, where $\mathcal{L}_b$ is the loss function for punishing undesirable logic rules. 
We do not include any soft priors in our current implementation.
\looseness=-1

\subsection{Consequent Estimation \texorpdfstring{$p(y | \bm{\alpha})$}{Lg}}
\label{sec:consequent_esimation}

Consequent estimation ensures a meaningful and consistent relationship between an explanation $\bm{\alpha}$ and prediction $y$, so each explanation naturally leads to the prediction according to human perception. 
This is achieved by testing the logic rule $\bm{\alpha} \Rightarrow y$ across the entire training dataset to ensure that it represents a global pattern that is typically consistent with human understanding.
\looseness=-1

\textbf{Empirical estimation}.
A straightforward way to compute $p(y | \bm{\alpha})$ is to first obtain all samples that satisfy antecedent $\bm{\alpha}$, and then calculate the percentage of them that have label $y$~\cite{peake2018explanation}.
For example, 
given explanation $\bm{\alpha}=$``\emph{awesome}$\geq$2'', if we obtain all instances in which \emph{awesome} appears more than twice and find that 90\% of them have label $y=$ \emph{positive sentiment}, then $p(y | \bm{\alpha})=0.9$.
Large $p(y | \bm{\alpha})$ corresponds to global patterns that naturally align with human perception.
Mathematically, this is equivalent to approximating $p(y | \bm{\alpha})$ with the empirical probability $\hat{p}(y| \bm{\alpha})$:
\looseness=-1
\begin{align}
\label{eq:empirical}
    \hat{p}(y| \bm{\alpha})
    = n_{\bm{\alpha},y} / n_{\bm{\alpha}}
\end{align}
where $n_{\bm{\alpha},y}$ is the number of training samples that satisfy the antecedent ${\bm{\alpha}}$ and has the consequent $y$, and $n_{\bm{\alpha}}$ is the number of training samples that satisfy the antecedent $\bm{\alpha}$.
\looseness=-1

Directly setting $p(y | \bm{\alpha})$ to $\hat{p}(y| \bm{\alpha})$ can cause two problems. First, when $n_{\bm{\alpha}}$ is not large enough, the empirical probability $\hat{p}(y| \bm{\alpha})$ may be inaccurate, and the modeling of such uncertainty is inherently missing in this formulation. 
Second, computing $\hat{p}(y| \bm{\alpha})$ for every antecedent $\bm{\alpha}$ is intractable, since the number of feasible antecedents $A$ increases exponentially with antecedent length $L$.
\looseness=-1
 
\textbf{Neural estimation of categorical distribution}.
To address the aforementioned problems, we jointly model $\hat{p}(y|\bm{\alpha})$ and the uncertainty caused by low-coverage antecedents with the categorical distribution and use a neural network to generalize to similar rules and better handle noise.
\looseness=-1

Assume that given antecedent $\bm{\alpha}$, $y$ follows a categorical distribution, with each category corresponding to a class. Then, according to the posterior predictive distribution, $y$ takes one of $K$ potential classes, and we may compute the probability of a new observation $y$ given existing observations:
\looseness=-1
\begin{align}
    \label{eq:categorical_distribution}
    p(y | \bm{\alpha}) 
    = p(y|\mathcal{Y}_{\bm{\alpha}}, \beta)
    \approx {{\hat{p}(y|\bm{\alpha})n_{\bm{\alpha}} + \beta } \over {n_{\bm{\alpha}} + K\beta}}
\end{align}
Here, $\mathcal{Y}_{\bm{\alpha}}$ denotes $n_{\bm{\alpha}}$ observations of class label $y$ obtained by checking the training data, and $\beta$ is the concentration hyperparameter of the categorical distribution that we automatically learn with backpropagation.
Eq.~(\ref{eq:categorical_distribution}) becomes Eq.~(\ref{eq:empirical}) when $n_{\bm{\alpha}}$ increases to $\infty$, and becomes a uniform distribution when $n_{\bm{\alpha}}$ goes to 0.
Thus, a low-coverage antecedent with a small $n_{\bm{\alpha}}$ is considered uncertain (i.e., close to uniform distribution).
By optimizing Eq.~(\ref{eq:categorical_distribution}), our method automatically balance the empirical probability $\hat{p}(y| \bm{\alpha})$ and the number of observations $n_{\bm{\alpha}}$. 
Probability $p(y|\bm{\alpha})$ also serves as the \textbf{confidence} score for the logic rule $\bm{\alpha}\Rightarrow y$.
\looseness=-1

We then employ a neural model to predict $\hat{p}(y|\bm{\alpha})$ and $n_{\bm{\alpha}}$ to better manage noise, generalize to similar rules, and improve efficiency. 
In particular, we obtain $A'$ samples of $\bm{\alpha}$ and compute $\hat{p}(y|\bm{\alpha})$ and $n_{\bm{\alpha}}$ by checking the training data. 
Here $A'$ is significantly smaller than the total number of feasible antecedents $A$ (Sec.~\ref{sec:optimization}).
We use the multi-task learning framework in~\cite{kendall2018multi} to train the neural network with these samples.
In particular, we minimize the loss in following equation.
\looseness=-1
\begin{align}
\label{eq:loss_c}
\mathcal{L}_c = {1 \over 2\sigma_p^2}||\hat{p}(y|\bm{\alpha})-\tilde{p}(y|\bm{\alpha})||^2 + 
    {1 \over 2\sigma_n^2}||n_{\bm{\alpha}}-\tilde{n}_{\bm{\alpha}}||^2 + \log \sigma_p \sigma_n
\end{align}
$\tilde{p}(y|\bm{\alpha})$, $\tilde{n}_{\bm{\alpha}}$ are the predicted empirical probability and the coverage given by the neural model, and $\sigma_p$ and $\sigma_n$ are standard deviations of ground truth probability and coverage.
More details for training the neural network are described in Appendix~\ref{sec:neural_consequent_estimation} and Appendix.~\ref{sec:differentiable_learning}, and effectiveness of the neural consequent estimator is shown in Appendix~\ref{sec:effectiveness_consequent_estimator}
\looseness=-1

\subsection{Deep Antecedent Generation \texorpdfstring{$p(\bm{\alpha} | \mathbf{x})$}{Lg}}
\label{sec:rule_generator} 
Deep antecedent generation finds explanation $\bm{\alpha}$ by reshaping the given deep model $f$.
Specifically, we replace the prediction layer in $f$ with an explanation generator, so that the latent representation $\mathbf{z}$ of input $\mathbf{x}$ is mapped to an explanation, instead of directly mapping to a prediction (e.g., class label).
\looseness=-1

Given $\mathbf{z}$, which is the representation of input $\mathbf{x}$ in the last hidden layer of $f$, we generate explanation $\bm{\alpha}=(o_1...,o_L)$ with a recursive formulation to ensure that the complexity is linear with $L$ (Sec.~\ref{sec:optimization}).
Formally, given $\mathbf{z}$ and $o_1,...o_{i-1}$, we obtain $o_i$ by
\looseness=-1
\begin{align}
    \label{eq:rule_prob}
    \centering
 \mathbf{h}_i = Encoder([\mathbf{z};\mathbf{o}_1...; \mathbf{o}_{i-1}]), \ \ \ \ \ 
 p(o_i|\mathbf{x}, o_1...,o_{i-1})  = {\mathbb{I}(o_i\in\mathcal{C}_i)  \text{exp}(\mathbf{h}_i^T \mathbf{o}_i) \over \sum_{\bm{\alpha}'_i} \mathbb{I}(\bm{\alpha}'_i\in\mathcal{C}_i) \text{exp}(\mathbf{h}_i^T \bm{\alpha}'_i )} 
\end{align}
where $\mathbf{o}_i$ is the embedding of $o_i$ and $Encoder$ is a neural sequence encoder such as GRU~\cite{cho2014properties} or Transformer~\cite{vaswani2017attention}.
$\mathbb{I}$ is the indicator function, and $\mathcal{C}_i$ is the set of candidates for $o_i$.
Every candidate should satisfy both global and local constraints.
The hard priors in Sec.~\ref{sec:prior} provide the global constraint and ensure that $\bm{\alpha}$ has a human-defined logic form. 
The local constraint requires that $\mathbf{x}$ satisfies antecedent $\bm{\alpha}$. An atom  ``\emph{awesome}$ \geq$ 2'', for example, is sampled only if $\mathbf{x}$ mentions ``\emph{awesome}'' more than once.\looseness=-1

We then sample $o_i$ from $p(o_i|\mathbf{x}, o_1...,o_{i-1})$ in a differentiable way to ensure easy end-to-end training:
\begin{align}
o_i = Gumbel(p(o'_i\in \mathcal{O} |\mathbf{x}, o_1...,o_{i-1})), \ \ \ \ \  p(\bm{\alpha}|\mathbf{x}) = \prod_{i\in[1,L]} p(o_i|\mathbf{x}, o_1...,o_{i-1})
\end{align}
$Gumbel$ is Straight-Through Gumbel-Softmax~\cite{jang2017categorical}, a differentiable function for sampling discrete values.
$o_i$ is represented as a one-hot vector with a dimension of $|\mathcal{O}|$ and is multiplied with the embedding matrix of atoms and logical connectives to derive the embedding $\mathbf{o}_i$.
\looseness=-1

\subsection{Optimization and Complexity Analysis}
\label{sec:optimization}

\textbf{Optimization}.
A deep logic rule reasoning model is learned in two steps. The first step optimizes the neural consequent estimator by minimizing loss $\mathcal{L}_c$ in Eq.~(\ref{eq:loss_c}). 
The neural consequent estimator only needs to be trained once for each dataset, and then it can be used for various deep models and hyperparameters. 
The second step converts deep model $f$ to an explainable version by maximizing $p(y|\mathbf{x},b)$ in Eq.~(\ref{eq:prob_approximation}) with a cross-entropy loss.
This is equivalent to minimizing loss $\mathcal{L}_d = - \mathcal{L}_b (\bm{\alpha}^{(s)})-\log\ p(y^*|\bm{\alpha}^{(s)})$, where $- \mathcal{L}_b (\bm{\alpha}^{(s)})$ punishes explanations that do not fit human's prior preference for rules (global coherency), and $\log\ p(y^*|\bm{\alpha}^{(s)})$ finds explanation $\bm{\alpha}^{(s)}$ that leads to the ground-truth class $y^*$ with a large confidence (prediction accuracy), in which the confidence is measured by testing rule $\bm{\alpha}^{(s)}\Rightarrow y^*$ in all training data (local coherency).
\looseness=-1

\begin{wraptable}[11]{h}{6.8cm}
	\centering
	\vspace{-4mm}
	\caption{
	Time complexity analysis. -RG and -NE denote our method without recursive antecedent generation and neural consequent estimator.
	}
	\resizebox{.5\columnwidth}{!}
	{
        \begin{tabular}{l|c|c}
        \hline
               & \begin{tabular}[c]{@{}c@{}}Consequent\\ Estimator\end{tabular} & \begin{tabular}[c]{@{}c@{}}Antecedent\\ Generator\end{tabular} \\ \hline
        SELOR  & $O(A'C+A'L^2)$                                                 & $O(N|\mathcal{O}|L+NL^2)$                                      \\
        -RG    & $O(A'C+A'L^2)$                                                 & $O(NA+NL^2)$                                                   \\
        -RG-NE & $O(AC)$                                                        & $O(NA)$                                                        \\ \hline
        \end{tabular}
	}
\label{tab:time-complexity}
\end{wraptable}
\textbf{Complexity analysis}.
Time complexity is compared in Table~\ref{tab:time-complexity}.
The complexity for antecedent generation corresponds to the time added for generating the antecedents during model training compared to the time required for training the base deep model $f$.
Here, $N$ is the number of training samples, and $C$ is the time complexity for computing the consequent of each antecedent.
As shown in the table, removing the recursive antecedent generator (RG) or the neural consequent estimator (NE) brings an additional linear complexity with the number of feasible antecedents $A$, which is much larger than $A'$. 
For example, in our experiment, setting $A'$ to $10^4$ is good enough to train an accurate neural consequent estimator, while the number of all possible antecedents is $A=6.25\times10^{12}$.
Here, we do not include the analysis for sampling $A'$ rules before training the consequent estimator. See Appendix~\ref{sec:neural_consequent_estimation} for more details.
\looseness=-1

%% file: source/5experiments.tex
\section{Experiment}
\label{sec:experiment}

\subsection{Experimental Settings}
\label{sec:experimental_setting}
\textbf{Datasets}.
We conduct experiments on three datasets.
The first two are textual, and the third is tabular.
\textbf{Yelp} classifies reviews of local businesses into positive or negative sentiment~\cite{zhang2015character}, and \textbf{Clickbait News Detection} from Kaggle labels whether a news article is a clickbait~\cite{clickbait2020kaggle}. 
\textbf{Adult} from the UCI machine learning repository~\cite{uci_ml_repository}, is an imbalanced tabular dataset that provides labels about whether the annual income of an adult is more than \$50K/yr or not.
For Yelp, we use a down-sampled subset (10\%) for training, as per existing work~\cite{okajima2019deep}.
More details about the datasets are in Appendix~\ref{sec:dataset}.
\looseness=-1

\textbf{Baselines}.
We compare our model to four baselines.
Two self-explainable models, \textbf{SENN}~\cite{alvarez2018towards} and \textbf{RCN}~\cite{okajima2019deep}, are compared in accuracy, robustness, explainability, and efficiency. 
Two post-hoc explainable methods, \textbf{LIME}~\cite{ribeiro2016should}and \textbf{Anchor}~\cite{ribeiro2018anchors}, are compared in explainability and efficiency.
\looseness=-1

\textbf{Implementation details}. 
To match with baselines, we use the AND operation by default in explanations.
The impact of using other logical connectives is presented in Appendix~\ref{sec:different_logical_connectives}.
The atoms, or interpretable features, are the same as in the majority of baselines, i.e., the existence of words for the textual dataset (e.g., ``\emph{amazing}''), and categorical and numerical features for the tabular data (e.g.,``\emph{age}<28'').
More details including selection of atom candidates are in Appendix~\ref{sec:implementation_details}.
\looseness=-1

\begin{table}[t]
\vspace{-3mm}
\centering
\caption{Comparison of classification performance measured in AUC. 
The average results from five runs are shown.
``Base'' refers to the performance of unexplainable vanilla backbones.
The best results among self-explaining models are marked in \textbf{bold}, and the \colorbox{blue!15}{highlighted cells} indicate a similar or better result compared with the unexplainable base model. The numbers in subscript indicates the standard error of the result.
}
\resizebox{0.9\textwidth}{!}
{
\begin{tabular}{lccccc|c}
\hline
 & \multicolumn{2}{c}{Yelp} & \multicolumn{2}{c}{Clickbait} & Adult &  \\ \cline{2-6}
 & BERT & \multicolumn{1}{c|}{RoBERTa} & BERT & \multicolumn{1}{c|}{RoBERTa} & DNN & \multirow{-2}{*}{Average} \\ \hline
Base & 97.39$_{\ 0.0659}$ & \multicolumn{1}{c|}{97.90$_{\ 0.0577}$} & 62.27$_{\ 1.0400}$ & \multicolumn{1}{c|}{63.72$_{\ 0.8722}$} & 68.62$_{\ 0.2317}$ & 77.98 \\ \hline
SENN & 96.00$_{\ 0.1087}$ & \multicolumn{1}{c|}{96.97$_{\ 0.0841}$} & 55.64$_{\ 1.0118}$ & \multicolumn{1}{c|}{57.93$_{\ 0.7779}$} & 67.39$_{\ 0.0854}$ & 63.20 \\
RCN & \cellcolor[HTML]{CBCEFB}\textbf{97.31$_{\ 0.0274}$} & \multicolumn{1}{c|}{\cellcolor[HTML]{CBCEFB}\textbf{98.03$_{\ 0.0086}$}} & 59.91$_{\ 0.2024}$ & \multicolumn{1}{c|}{59.37$_{\ 0.2259}$} & \cellcolor[HTML]{CBCEFB}\textbf{70.06$_{\ 0.0411}$} & 76.94 \\ \hline
SELOR & \cellcolor[HTML]{CBCEFB}\textbf{97.28$_{\ 0.0335}$} & \multicolumn{1}{c|}{\cellcolor[HTML]{CBCEFB}\textbf{97.78$_{\ 0.0833}$}} & \textbf{60.31$_{\ 0.8498}$} & \multicolumn{1}{c|}{\cellcolor[HTML]{CBCEFB}\textbf{64.14$_{\ 0.5906}$}} & \cellcolor[HTML]{CBCEFB}\textbf{70.36$_{\ 0.0892}$} & \cellcolor[HTML]{CBCEFB}\textbf{77.97} \\ \hline
\end{tabular}
}
\label{tab:prediction_result}
\end{table}

\subsection{Classification Performance and Robustness}

\textbf{Classification performance}.
Table~\ref{tab:prediction_result} shows the classification performance of \pname{} and baselines.
Here, we evaluate the PR AUC instead of the ROC AUC because the latter is less suitable for imbalanced datasets~\cite{saito2015precision}.
BERT~\cite{kenton2019bert} and RoBERTa~\cite{liu2019roberta} are used as the backbone networks in the NLP datasets, while 3-Layer DNN is used for the tabular dataset. 
The base method is the vanilla backbone network that does not support explainability (Appendix~\ref{sec:implementation_details}). The prediction performance of post-hoc methods, LIME and Anchor, is the same as the base model as they utilize the trained model without any extra optimization. Comparison with a fully-transparent model is presented in Appendix~\ref{sec:fully_transparent}.
Our method achieves comparable average performance with the unexplainable base model and outperforms other self-explaining models by 1.3\%.
Moreover, our method achieves the best or comparable results on various datasets against backbone models, demonstrating the expressive power of logic rules for high prediction performance.
RCN cannot perform as well on challenging textual datasets like Clickbait because it computes soft attention over a predefined rule set.
This indicates that (potentially biased) predefined rule sets will limit the model's capability.
\looseness=-1  

\begin{figure}[t]
\vspace{-2mm}
    \centering
    \begin{subfigure}[t]{0.31\textwidth}
        \centering
        \includegraphics[width=\textwidth]{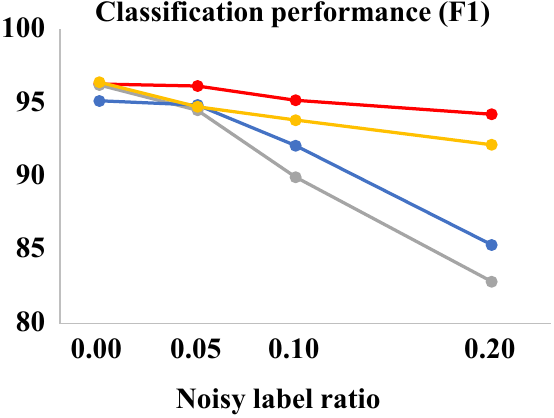}
        \caption{Yelp}
        \label{fig:robustness_yelp}
    \end{subfigure}
    \hspace{1mm}
    \begin{subfigure}[t]{0.31\textwidth}
        \centering
        \includegraphics[width=\textwidth]{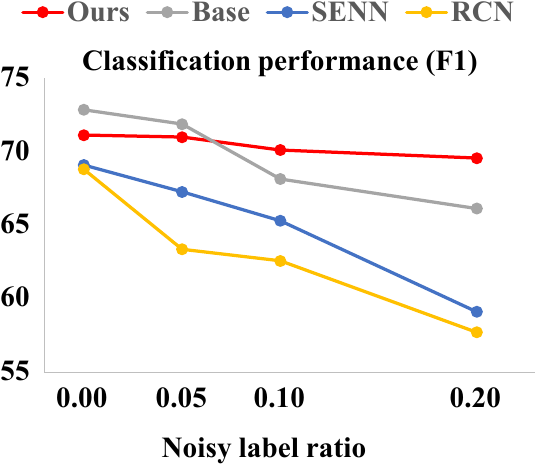}
        \caption{Clickbait}
        \label{fig:robustness_clickbait}
    \end{subfigure}
    \hspace{1mm}
    \begin{subfigure}[t]{0.31\textwidth}
        \centering
        \includegraphics[width=\textwidth]{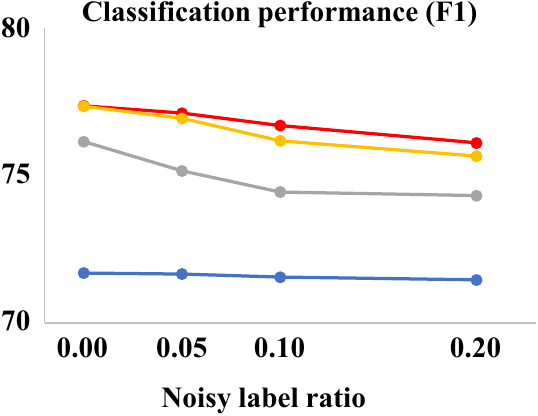}
        \caption{Adult}
        \label{fig:robustness_adult}
    \end{subfigure}
    \caption{Experimental results on robustness under different ratios of noisy labels.\looseness=-1}
    \vspace{-6mm}
    \label{fig:robustness}
\end{figure}

\textbf{Robustness to noisy labels}. 
Following the literature~\cite{li2019dividemix,zhang2018mixup}, we assess the robustness of \pname{} against randomly corrupted labels. We hypothesize that the effect of the noisy label is alleviated by consequent estimation term $p(y|\bm{\alpha})$, where model verifies its decision by testing the logic rule over the entire dataset.
For experiments, symmetric noise is introduced by randomly flipping the labels for a subset of the training data.
Fig.~\ref{fig:robustness} shows the results over Yelp, Clickbait, and Adult datasets with multiple levels of noise ratio from 5\% to 20\%. 
Our model outperforms other models across all scenarios. 
The improvement is substantial even with a high noise ratio (i.e., 20\%).
For a noise ratio above 10\%, our method consistently outperforms the unexplainable base models (2.4\% to 13.7\%).
\looseness=-1

\textbf{Sensitivity analysis.}
Due to space limitations, we show that the prediction performance of SELOR is stable under different hyper-parameter settings in Appendix~\ref{sec:hyperparameter_sensitivity}.

\subsection{Explainability}
\label{sec:explainability}

\begin{figure}[b]
\vspace{-4mm}
\includegraphics[width=1\textwidth]{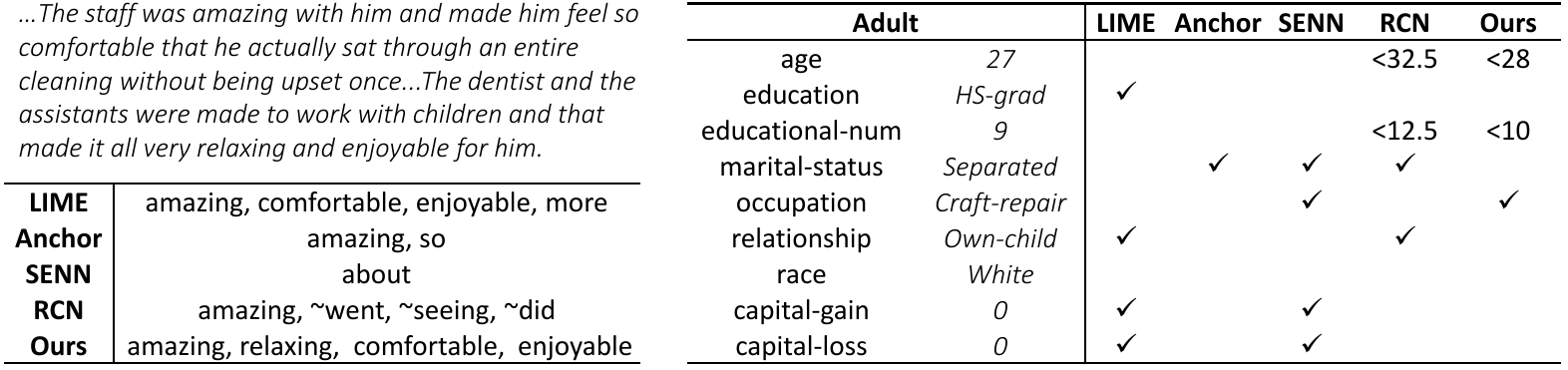}
\caption{Example explanations produced by five methods on Yelp (left) and Adult (right).}
\label{fig:cases}
\end{figure}

\textbf{User study on human precision}.  
To evaluate human precision, we recruited nine native English speakers through a vendor company~\cite{speechocean}.
Each participant was provided with randomly selected 50 Yelp reviews and 50 Adult samples.
Five explanations obtained from different methods were provided for each sample, and the participants reviewed whether the explanations offered reasonable rationales. 
Participants provided two labels for each explanation, indicating whether it was good or if it was the best one.
A \textbf{good} explanation should naturally lead to the prediction, but it can contain noisy features.
For example, ``\emph{amazing}, \emph{are}'' is a good explanation for positive sentiment.
The \textbf{best} explanation is the one that contains the most important and least noisy features.
The participants were allowed to choose multiple best explanations only if the chosen ones were the same.
For a fair comparison, we showed explanations in the same form: a list of features each method considers important for prediction.
Example explanations generated by our method and the baselines are shown in Fig.~\ref{fig:cases}. 
Note that \~{} in RCN means negation.
More details about the explanation generation, labeling guidelines, and participants' results are given in Appendix~\ref{sec:user_study_2_detail}.
\looseness=-1

Table~\ref{tab:user_study} shows that \pname{} marks the highest percentage of good explanations, with an average ratio of 94.4\% on Yelp and 90.7\% on Adult.
Our method is also most frequently chosen as the best explanation. 
All results are statistically significant according to the p-values from the t-tests.
Although logic rules are promising, choosing from a small set of predefined rules may be insufficient due to the potential bias in the rule set.
For example, RCN uses rules extracted with traditional machine learning methods that meet the global data distribution but frequently fail to adequately represent each sample, particularly on datasets with many features like Yelp.
As a result, RCN is rarely chosen as the best explanation, especially for Yelp text data.
Post-hoc methods also tend to offer good human precision.
The best ratio of LIME and Anchor, however, is substantially lower than ours, indicating that the base model may rely on more noisy features for prediction. In contrast, our method can verify its decision by testing the logic rule across the entire dataset.
\looseness=-1

\begin{table}[t]
\centering
\caption{
User study results on human precision.
We show the average (Avg.) and inter-participant agreement (Agr.) on the percentage of explanations that are considered good (a, b) or best (c, d).
One star (*) means p-value is less than 0.05.
Best results are highlighted in \textbf{bold}.\looseness=-1
}
\vspace{-1mm}
\begin{subtable}[h]{0.38\textwidth}
\caption{Percentage of good (Yelp)}
\vspace{-2mm}
\resizebox{\textwidth}{!}
{
\begin{tabular}{l|c|c|c}
\hline
 & Avg. & Agr. & P-value \\ \hline
Lime & 89.8 & 84.4 & 8.68 E-04* \\
Anchor & 84.4 & 87.7 & 1.12 E-07* \\
SENN & 34.4 & 72.3 & 1.40 E-51* \\
RCN & 64.0 & 77.6 & 7.26 E-13* \\ \hline
SELOR & \textbf{94.4} & 93.9 & - \\ \hline
\end{tabular}
}
\label{tab:good_yelp}
\end{subtable}
\begin{subtable}[h]{0.38\textwidth}
\caption{Percentage of good (Adult)}
\vspace{-2mm}
\resizebox{\textwidth}{!}
{
\begin{tabular}{l|c|c|c}
\hline
 & Avg. & Agr. & P-value \\ \hline
Lime & 42.7 & 57.1 & 6.09 E-54* \\
Anchor & 52.7 & 59.9 & 5.56 E-18* \\
SENN & 46.0 & 51.5 & 1.18 E-41* \\
RCN & 60.9 & 53.2 & 2.83 E-27* \\ \hline
SELOR & \textbf{90.7} & 85.7 & - \\ \hline
\end{tabular}
}
\label{tab:good_adult}
\end{subtable}
\begin{subtable}[h]{0.38\textwidth}
\vspace{-2mm}
\caption{Percentage of best (Yelp)}
\vspace{-2mm}
\resizebox{\textwidth}{!}
{
\begin{tabular}{l|c|c|c}
\hline
 & Avg. & Agr. & P-value \\ \hline
Lime & 34.2 & 67.6 & 8.87 E-03* \\
Anchor & 18.0 & 83.6 & 5.63 E-18* \\
SENN & 2.4 & 96.3 & 6.84 E-40* \\
RCN & 2.0 & 96.3 & 6.84 E-40* \\ \hline
SELOR & \textbf{46.7} & 64.8 & - \\ \hline
\end{tabular}
}
\label{tab:best_yelp}
\end{subtable}
\begin{subtable}[h]{0.38\textwidth}
\vspace{-2mm}
\caption{Percentage of best (Adult)}
\vspace{-2mm}
\resizebox{\textwidth}{!}
{
\begin{tabular}{l|c|c|c}
\hline
 & Avg. & Agr. & P-value \\ \hline
Lime & 1.3 & 96.7 & 1.72 E-64* \\
Anchor & 13.8 & 82.9 & 1.23 E-35* \\
SENN & 9.6 & 83.3 & 2.30 E-36* \\
RCN & 10.2 & 82.9 & 1.23 E-35* \\ \hline
SELOR & \textbf{65.1} & 58.4 & - \\ \hline
\end{tabular}
}
\label{tab:best_adult}
\end{subtable}
\vspace{-4mm}
\label{tab:user_study}
\end{table}

\textbf{Case study on model debugging and refinement}.
What useful insights can \pname{} provide on performance? 
In a study of 20,000 sampled Yelp reviews, we clustered the generated explanations into 10 clusters by applying K-Means on the antecedent embeddings.
Table~\ref{tab:case-study} shows five clusters with the lowest training accuracy to illustrate \reason{potential reasons} for  bad performance.
Here, NULL is an empty atom when the model generated explanations that were shorter than the predefined length $L$.
\looseness=-1

\begin{table}[t]
	\centering
	\caption{
    Case study on Yelp. 
    We cluster the explanations for the training samples and show the five clusters with the lowest training accuracy. Num is the number of explanations in the cluster, and Len is the average text length of the reviews. 
    Potential reasons for bad performance are marked in \reason{brown}. 
	}
	\resizebox{\columnwidth}{!}
	{
        \begin{tabular}{c|c|c|c|c|c}
        \hline
        Cluster & Acc & Label & Num & Len & Atoms in the explanations (ordered by frequency) \\ \hline
        1 & 99.2 & \begin{tabular}[c]{@{}c@{}}99.1\%\\ Neg\end{tabular} & \begin{tabular}[c]{@{}c@{}}1,763\\ 8.82\%\end{tabular} & 643 & \begin{tabular}[c]{@{}c@{}}not(290) bad(233) no(185) mediocre(153) bland(149) never(123)\\ again(122) worst(119) ok(115) disappointing(115) terrible(107)\end{tabular} \\ \hline
        2 & 99.2 & \begin{tabular}[c]{@{}c@{}}99.2\%\\ Pos\end{tabular} & \begin{tabular}[c]{@{}c@{}}2,730\\ 13.7\%\end{tabular} & 584 & \begin{tabular}[c]{@{}c@{}}great(667) delicious(508) best(330) love(294) fresh(285) tasty(255)\\ definitely(254) friendly(232) perfect(199) amazing(198) favorite(194)\end{tabular} \\ \hline
        3 & 98.6 & \begin{tabular}[c]{@{}c@{}}98.6\%\\ Pos\end{tabular} & \begin{tabular}[c]{@{}c@{}}2,686\\ 13.4\%\end{tabular} & 548 & \begin{tabular}[c]{@{}c@{}}great(793) friendly(329) always(315) best(304) love(295) fun(223)\\ definitely(218) helpful(163) awesome(159) amazing(136) \reason{vegas(117)}\end{tabular} \\ \hline
        4 & 93.2 & \begin{tabular}[c]{@{}c@{}}57.8\%\\ Pos\end{tabular} & \begin{tabular}[c]{@{}c@{}}848\\ 4.24\%\end{tabular} & \reason{119} & \begin{tabular}[c]{@{}c@{}}\reason{NULL(1738)} great(133) not(67) best(39) service(39) love(34)\\ friendly(29) good(26) awesome(22) fast(22) overpriced(20)\end{tabular} \\ \hline
        5 & 83.9 & \begin{tabular}[c]{@{}c@{}}58.1\%\\ Neg\end{tabular} & \begin{tabular}[c]{@{}c@{}}62\\ \reason{0.31\%}\end{tabular} & 682 & \begin{tabular}[c]{@{}c@{}}\reason{NULL(24) nicht(13) un(11) eine(9)} service(8) \reason{pas(8) der(8)} die(7)\\ \reason{und(7) um(6) den(5) pour(5) de(5) das(4) prix(4) je(4) zu(3) im(3)}\end{tabular} \\ \hline
        \end{tabular}
	}

\label{tab:case-study}
\end{table}

We make the following observations.
First, low training accuracy in \textbf{cluster 5} is due to non-English reviews, which accounted for 0.31\% and led to underfitting.
Second, performance degradation also happens when the model does not have enough evidence. For example, reviews in \textbf{cluster 4} were short (average length of 119 words) and contained an overwhelming number of NULL atoms (on average 2 per explanation).
Third, \textbf{cluster 3} contained 13.4\% samples with positive sentiment, and its training accuracy (98.6\%) is higher than cluster 4. However, the cluster often included ``\emph{vegas}'' in the explanation, which does not seem directly related to sentiment classification.
Fourth, \textbf{clusters 1 and 2} have reasonable atoms, which seem consistent with high training accuracy (99.2\%).
\looseness=-1

\begin{figure}[t]
    \centering
    \includegraphics[width=.95\textwidth]{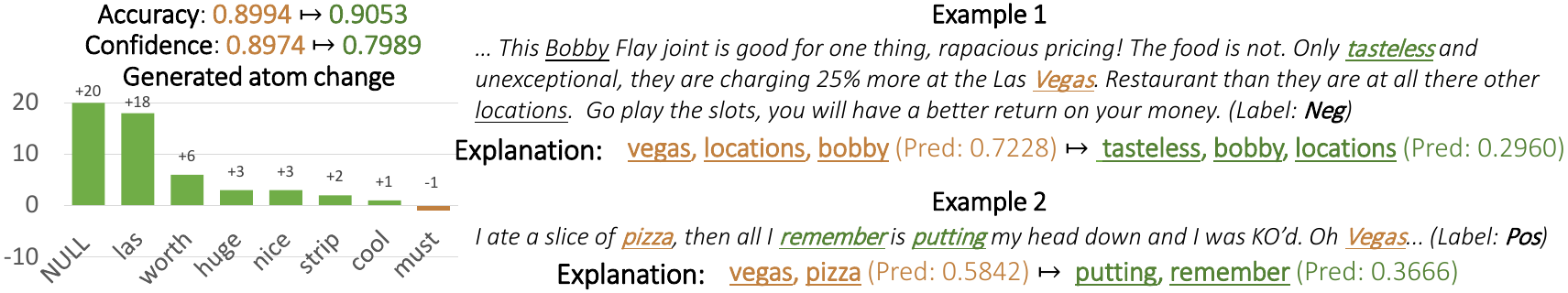}
    \vspace{-1mm}
    \caption{Steering the model without re-training.
    \pname{} allows users to exclude noisy features from explanations during testing, which may simultaneously improve the explanation quality and prediction accuracy. This figure shows the performance \reason{before} and \after{after} removing  ``\emph{vegas}''. \looseness=-1
    }
    \label{fig:vegas}
\end{figure}

\pname{} allows us to steer the model directly. 
For example, after identifying the potentially noisy feature like ``\emph{vegas}'', we can prevent the model from including the term by removing it from the candidate atom list $\mathcal{C}_i$. 
This type of refinement can be easily achieved during testing, unlike the efforts-taking dataset calibration or model retraining.
Fig.~\ref{fig:vegas} shows the performance change of the 169 samples that previously included ``\emph{vegas}'' in their explanations.  
The histogram shows which atoms that are generated more after removing ``\emph{vegas}''.
The model sometimes relies on similar atoms such as ``\emph{las}'' or does not find a good candidate (e.g., choosing NULL), which may lead to decreased confidence. 
However, the chance of including more meaningful atoms also increases (e.g., ``\emph{worth}'' in the histogram, and ``\emph{tasteless}'' in Example 1). 
One may also verify assumptions by checking the samples whose prediction score changes.
For instance, after removing ``\emph{vegas}'', the model can no longer predict Example 2 correctly.
The example contains no obvious indication of sentiment, and ``\emph{vegas}'' may be the most helpful feature.
This contradicts our previous assumption that ``\emph{vegas}'' seems not critical for sentiment classification. Instead it can provide new insights and guidance for further improvement (e.g., punishing ``\emph{vegas}'' with a soft prior instead of directly removing it). \looseness=-1 

\textbf{Explanation stability and sensitivity analysis}.
We discuss the stability of our explanations in Appendix~\ref{sec:explanation_stability}.
Our quantitative experiment demonstrates that the explanations generated in different runs are consistent. We also present a case study in that SELOR gives similar explanations for similar inputs. Moreover, we discuss user study results that the human precision of the explanations is good across different hyper-parameter settings in Appendix~\ref{sec:hyperparameter_sensitivity}.
\looseness=-1

\subsection{Efficiency}
\label{sec:efficiency}
Table~\ref{tab:efficiency} shows that post-hoc explanation methods like LIME and Anchor require a longer time to generate an explanation. RCN has the largest complexity among the self-explaining methods since it enumerates all possible rules and combines them with soft attention.
To alleviate this problem, RCN uses a predefined rule set; hence, its efficiency becomes dependent on the size and quality of the rule set. 
In contrast, \pname{} is trained within acceptable time even for large solution space, and humans only need to define the types of atoms and logical connectives.
%
Our model generated each explanation with a linear complexity with length $L$, while RCN goes over all possible rules and has exponential complexity with $L$.
Our method required additional time for the neural consequent estimator, taking 35 minutes on Yelp and 25 minutes on Adult. This step is only required once for each dataset and hence is acceptable. The consequent estimator can also be reused.
\looseness=-1

\begin{table}[t]
\vspace{-2mm}
\centering
\caption{Time costs in seconds on Yelp (BERT) and Adult.}
\resizebox{1\textwidth}{!}
{
\begin{tabular}{l|c|cccc|ccccc}
\hline
 & \multirow{2}{*}{\begin{tabular}[c]{@{}c@{}}Consequent\\ estimator training\end{tabular}} & \multicolumn{4}{c|}{Deep model training (1 epoch)} & \multicolumn{5}{c}{Explanation generation (1 sample)} \\ \cline{3-11} 
 &  & Base & SENN & RCN & SELOR & LIME & Anchor & SENN & RCN & SELOR \\ \hline
Yelp & 2041.4 & 571.9 & 224.5 & 503.7 & 665.6 & 55.0 & 2854.2 & 0.037 & 0.071 & 0.055 \\ \hline
Adult & 1502.8 & 12.8 & 9.1 & 953.9 & 98.2 & 2.5 & 1.18 & 0.02 & 0.17 & 0.015 \\ \hline
\end{tabular}
}
\label{tab:efficiency}
\vspace{-5mm}
\end{table}

%% file: source/6conclusion.tex
\section{Conclusion and Future Work}
\label{sec:conclusion}

This work presented a new framework, \pname{}, which incorporates self-explanatory capabilities into a deep model to provide high human precision by explaining logic rules while also maintaining high prediction performance.
Our method does not require predefined rule sets and can be learned in a differentiable way.
Extensive tests involving human evaluation show that our method achieves high prediction performance and human precision while being resistant to noisy labels. 
Although our method brings substantial advantages, there remain multiple aspects for improvement in the future:\looseness=-1

\textbf{Stability}.
A desirable property for self-explaining models is stability, which requires that similar inputs lead to similar explanations.
Unlike SENN~\cite{alvarez2018towards}, which proposes a robustness loss to ensure stable explanations against adversarial inputs, \pname{} does not employ such a constraint and cannot guarantee the stability of explanations for inputs with similar raw features.
However, our framework theoretically ensures stability is modeled in the selected feature space (see Appendix~\ref{sec:analysis_explanation_stability} for more details), which is partially evaluated by a case study in Appendix~\ref{sec:explanation_stability}.
\looseness=-1

\textbf{Applicability}.
While we explored text and tabular data, our model is applicable to other data types like images and graphs. 
We can treat a cluster of images or superpixels as an atom~\cite{alvarez2018towards} or extract atoms with CAV (Concept Activation Vector), a feature that indicates the concept of humans (e.g., striped, red)~\cite{kim2018interpretability}. End-to-end feature learning is possible in our framework if the number of candidate atoms is small (e.g., around 100 object classes or concepts~\cite{yu2022probabilistic}). 
\looseness=-1


\textbf{Level of insight}.
\pname{} cannot explicitly model higher-level properties of atoms (e.g., learn that ``\emph{awesome}'' is a \emph{positive sentiment word} and make a rule based on \emph{positive sentiment word}) since we do not directly consider predicates. We can only find rules constructed with bottom-level atoms instead of summarizing important high-level patterns, which also leads low coverage of rules (e.g., rule ``\emph{awesome} AND \emph{tasty}'' => \emph{positive sentiment} only covers 0.37\% of the input instances).
If re-designed as the first-order logic, the model may directly find high-level patterns such as ``\emph{a negation word} AND \emph{a positive sentiment word}'' => \emph{negative sentiment}, instead of listing many specific rules such as ``\emph{not great}'' => \emph{negative sentiment} and ``\emph{no good}'' => \emph{negative sentiment}.
This could save human cognitive budget and improve the reasoning capability of deep models.
Moreover, we may automatically compose high-level concepts such as ``\emph{strong positive phrase}'' and build rules with them.
The concept ``\emph{strong positive phrase}'' may be composed by detecting two consecutive positive sentiment words (``\emph{amazingly comfortable}'' and ``\emph{perfectly enjoyable}'') with predicate invention in~\cite{evans2018learning}.
\looseness=-1

%% file: source/acknowledgement.tex
\begin{ack}
We thank Fangzhao Wu, Sundong Kim, and Eunji Lee for their insightful feedback on our work. We appreciate the reviewers of this paper for their valuable suggestions that improved the paper significantly. This research was supported by Microsoft Research Asia, the Institute for Basic Science (IBS-R029-C2) in Korea, and the Potential Individuals Global Training Program (2021-0-01696) by the Ministry of Science and ICT in Korea.
\end{ack}

%% file: source/checklist.tex
\section*{Checklist}

\begin{enumerate}

\item For all authors...
\begin{enumerate}
  \item Do the main claims made in the abstract and introduction accurately reflect the paper's contributions and scope?
    \answerYes{}
  \item Did you describe the limitations of your work?
    \answerYes{} See Sec.~\ref{sec:conclusion}.
  \item Did you discuss any potential negative societal impacts of your work?
    \answerYes{} See Sec.~\ref{sec:conclusion}.
  \item Have you read the ethics review guidelines and ensured that your paper conforms to them?
    \answerYes{}
\end{enumerate}

\item If you are including theoretical results...
\begin{enumerate}
  \item Did you state the full set of assumptions of all theoretical results?
    \answerYes{} See Sec.~\ref{sec:logic_formulation}, Sec.~\ref{sec:framework_logic}, Sec.~\ref{sec:prior} and Appendix~\ref{sec:probability_decomposition}.
  \item Did you include complete proofs of all theoretical results?
    \answerYes{} See Section~\ref{sec:framework_logic}, Section~\ref{sec:consequent_esimation} and Appendix~\ref{sec:probability_decomposition}
\end{enumerate}

\item If you ran experiments...
\begin{enumerate}
  \item Did you include the code, data, and instructions needed to reproduce the main experimental results (either in the supplemental material or as a URL)?
    \answerYes{} Codes are released at Github (\emph{https://github.com/archon159/SELOR}).
  \item Did you specify all the training details (e.g., data splits, hyperparameters, how they were chosen)?
    \answerYes{} See Appendix~\ref{sec:dataset} and Appendix~\ref{sec:implementation_details}.
  \item Did you report error bars (e.g., with respect to the random seed after running experiments multiple times)?
    \answerYes{} See Table~\ref{tab:prediction_result} and Table~\ref{tab:prediction_f1}.
  \item Did you include the total amount of compute and the type of resources used (e.g., type of GPUs, internal cluster, or cloud provider)?
    \answerYes{} See Appendix~\ref{sec:implementation_details}. 
\end{enumerate}

\item If you are using existing assets (e.g., code, data, models) or curating/releasing new assets...
\begin{enumerate}
  \item If your work uses existing assets, did you cite the creators?
    \answerYes{} See Sec.~\ref{sec:experimental_setting} and Appendix~\ref{sec:implementation_details}.
  \item Did you mention the license of the assets?
    \answerNA{}
  \item Did you include any new assets either in the supplemental material or as a URL?
    \answerYes{} See checklist 3-(a).
  \item Did you discuss whether and how consent was obtained from people whose data you're using/curating?
    \answerNA{}
  \item Did you discuss whether the data you are using/curating contains personally identifiable information or offensive content?
    \answerNA{}
\end{enumerate}

\item If you used crowdsourcing or conducted research with human subjects...
\begin{enumerate}
  \item Did you include the full text of instructions given to participants and screenshots, if applicable?
    \answerYes{} 
    See Sec.~\ref{sec:explainability}, Appendix~\ref{sec:intro_sup}, Appendix~\ref{sec:user_study_2_detail} and additionally attached guideline files and screenshots.
  \item Did you describe any potential participant risks, with links to Institutional Review Board (IRB) approvals, if applicable?
    \answerNA{}{}
  \item Did you include the estimated hourly wage paid to participants and the total amount spent on participant compensation?
    \answerYes{} See Appendix~\ref{sec:intro_sup} and Appendix~\ref{sec:user_study_2_detail}.
\end{enumerate}

\end{enumerate}

%% file: source/appendix.tex
\appendix
{\Large \textbf{Appendix}}

\input{source/appendix0_introduction}
\newpage
\input{source/appendix1_design}
\newpage
\input{source/appendix2_experiment}

%% file: source/appendix0_introduction.tex
\section{Supplement for Section 1 (Introduction)}
\label{sec:intro_sup}
Here, we present details of the user study in Fig.~\ref{fig:rule_human_precision}.
The figure shows that logic rule explanations achieve higher human precision than linear-regression-based explanations with local stability, while providing a confidence score that correlates with human precision. 

We used a vendor company to recruit three native English speakers for the user study (Sec.~\ref{sec:explainability}).
User studies can be performed by 1) hiring a large number of labelers from platforms like Prolific and AMT or 2) hiring a limited number of experienced annotators from a labeling company. While platforms like Prolific make it easy to find many labelers, they are known to be better suited for cognitively simple tasks and may suffer from errors~\cite{peer2022data}. Our task is challenging for ordinary labelers, as we require them to carefully reason about which features of an adult are useful for predicting his or her income (the Adult dataset) and compare multiple similar explanations. Thus, we validated the model with more experienced annotators hired through a labeling company. To ensure the labelers have an adequate understanding of the task, we provided them with detailed guidelines and examined their initial labels with feedback when a misunderstanding is detected. Such a close interaction would not be possible in crowdsourcing platforms, which may lead to errors and unreliable results.

Each participant was provided 1,000 and 500 randomly selected explanations from \pname{} and SENN, respectively.
For each explanation, we test whether it can naturally lead to the model prediction according to human perception.
Participants were asked to provide 1) the class label for the explanation and 2) how confident they were in their decision by using a 5-point likert scale (HC, i.e., human confidence).
For example, given an explanation ``\emph{awesome}, \emph{tasty}'', the participant will give the label \emph{positive sentiment} and a high confidence score ``5'' out of 5.
When labels were the same to model predictions, human precision was high.
We sampled explanations so that their confidence score from models (MC, i.e., model confidence) was evenly distributed and examined how explanation quality varies with the confidence score.
Fig.~\ref{fig:rule_human_precision} shows how human precision changes with different levels of model confidence. 
As shown in the figure, logic rule explanations achieve higher human precision than the linear-regression-based explanations, and the model confidence shows a strong correlation with human precision.
Here, human precision is the F1-score of machine prediction for give logic rules using human prediction as the ground-truth labels.
Table~\ref{tab:rule_human_precision} provides more detailed information about our user study. 
The logic rule with a higher MC level tends to have higher agreement and HC. Also, the logic rule shows better human precision at most MC levels.

\textbf{User instruction and labeling detail}.
We describe the instructions given to participants in the attached guideline file (\emph{Labeling\_Guidelines\_User\_Study\_Figure1b.pdf}) with detailed description of the task and labeling examples. 
Participants received an Excel file containing blank labels, which they were instructed to fill out and return. The snapshot of the Excel file is also attached as a separate file (\emph{Screenshot\_User\_Study\_Figure1b.PNG}). Each participant was paid $22.5\$$ per hour and the total budget we spent was $937.5\$$ for this task.

\begin{table}[b]
\vspace{-10mm}
\centering
\caption{User study results for human precision on logic rule- and linear-regression-based explanations. HC denotes the average human confidence while MC denotes the machine confidence. Avg. denotes the average number of sentiment agreement, human confidence, and human precision of all data points. (Lv 1: 0.0 $\sim$ 0.2, Lv 2: 0.2 $\sim$ 0.4, Lv 3: 0.4 $\sim$ 0.6, Lv 4: 0.6 $\sim$ 0.8, Lv 5: 0.8 $\sim$ 1.0)}

\begin{subtable}[h]{0.48\textwidth}
\caption{Logic rule}
\resizebox{\textwidth}{!}
{
\begin{tabular}{l|ccc}
\hline
 & \begin{tabular}[c]{@{}c@{}}Sentiment\\ Agreement\end{tabular} & Avg HC & \begin{tabular}[c]{@{}c@{}}Human\\ Precision\end{tabular} \\ \hline
MC Lv 1 & 82.67 & 2.72 & 52.65 \\
MC Lv 2 & 86.00 & 2.88 & 53.53 \\
MC Lv 3 & 84.00 & 3.31 & 76.19 \\
MC Lv 4 & 92.67 & 3.85 & 89.38 \\
MC Lv 5 & 95.00 & 4.07 & 90.41 \\ \hline
Avg. & 88.07 & 3.36 & 73.32 \\ \hline
\end{tabular}
}
\end{subtable}
\hspace{0.3mm}
\begin{subtable}[h]{0.48\textwidth}
\caption{Linear-regression-based explanation}
\resizebox{\textwidth}{!}
{
\begin{tabular}{l|ccc}
\hline
 & \begin{tabular}[c]{@{}c@{}}Sentiment\\ Agreement\end{tabular} & Avg HC & \begin{tabular}[c]{@{}c@{}}Human\\ Precision\end{tabular} \\ \hline
MC Lv 1 & 78.79 & 3.63 & 47.71 \\
MC Lv 2 & 81.56 & 3.55 & 48.69 \\
MC Lv 3 & 75.95 & 3.62 & 52.83 \\
MC Lv 4 & 79.12 & 3.56 & 56.82 \\
MC Lv 5 & 84.51 & 3.78 & 66.17 \\ \hline
Avg. & 79.96 & 3.63 & 54.46 \\ \hline
\end{tabular}
}
\end{subtable}
\label{tab:rule_human_precision}
\end{table}

%% file: source/appendix1_design.tex
\section{Supplement for Section 2 (Deep Logic Rule Reasoning)}
\subsection{Symbols}
\label{sec:symbol}
Table~\ref{tab:symbol} summarizes the symbols used in this paper.

\begin{longtable}{clm{0.6\textwidth}}
\caption{The meaning and detailed explanation of each symbol used in the paper.}
\label{tab:symbol} \\

\hline
 & Meaning & Detailed Explanation \\ \hline
\multicolumn{1}{c|}{$\mathbf{x}$} & \multicolumn{1}{l|}{Input sample} & Any type of data (e.g. text, tabular) \\ \hline
\multicolumn{1}{c|}{$\bm{\alpha}$} & \multicolumn{1}{l|}{Antecedent} & Condition to apply the rule \\ \hline
\multicolumn{1}{c|}{$b$} & \multicolumn{1}{l|}{Human belief} & Common sense that a human believes when they make a decision \\ \hline
\multicolumn{1}{c|}{$y$} & \multicolumn{1}{l|}{Consequent} & Model's prediction output for the given antecedent \\ \hline
\multicolumn{1}{c|}{\multirow{2}{*}{$o_i$}} & \multicolumn{1}{l|}{\multirow{2}{*}{Atom or logical connective}} & Atom is the smallest unit of explanation \\ \cline{3-3} 
\multicolumn{1}{c|}{} & \multicolumn{1}{l|}{} & Logical connective combines atoms \\ \hline
\multicolumn{1}{c|}{$\mathbf{o}_i$} & \multicolumn{1}{l|}{Embedding of $o_i$} & Initialized as the average embedding of all training samples that satisfy the atom \\ \hline
\multicolumn{1}{c|}{$\mathcal{O}$} & \multicolumn{1}{l|}{Set of atoms} &  \\ \hline
\multicolumn{1}{c|}{$\mathcal{C}_i$} & \multicolumn{1}{l|}{Set of candidates for $o_i$} & Every candidate should satisfy both global and local constraints (Sec.~\ref{sec:rule_generator}) \\ \hline
\multicolumn{1}{c|}{$L$} & \multicolumn{1}{l|}{Length of an antecedent} & Number of atoms and logical connectives included in an antecedent \\ \hline
\multicolumn{1}{c|}{$N$} & \multicolumn{1}{l|}{Number of training data} &  \\ \hline
\multicolumn{1}{c|}{$n_{\bm{\alpha}}$} & \multicolumn{1}{l|}{See detailed explanation} & Number of data samples in training data that satisfies the antecedent $\bm{\alpha}$ \\ \hline
\multicolumn{1}{c|}{$n_{\bm{\alpha}, y}$} & \multicolumn{1}{l|}{See detailed explanation} & Number of data samples in training data that satisfies the antecedent $\bm{\alpha}$ and has the consequent y\\ \hline
\multicolumn{1}{c|}{$\mathcal{Y}_{\bm{\alpha}}$} & \multicolumn{1}{l|}{See detailed explanation} & Data samples of class $y$ in training data that satisfies the antecedent $\bm{\alpha}$ \\ \hline
\multicolumn{1}{c|}{$\bm{\alpha}^{(s)}$} & \multicolumn{1}{l|}{s-th sample of $\bm{\alpha}$} & s-th sampled antecedent in deep antecedent generation \\ \hline
\multicolumn{1}{c|}{$S$} & \multicolumn{1}{l|}{Total number of $\bm{\alpha}^{(s)}$} & Set as $S$ = 1 by default \\ \hline
\multicolumn{1}{c|}{$\Omega(\bm{\alpha})$} & \multicolumn{1}{l|}{Required number of $\bm{\alpha}$} & The number of explanation required to explain given input \\ \hline
\multicolumn{1}{c|}{$\mathbf{h}_i$} & \multicolumn{1}{l|}{Hidden state of encoder} & The encoder can be any neural sequence encoder such as GRU or Transformer \\ \hline
\multicolumn{1}{c|}{$C$} & \multicolumn{1}{l|}{See detailed explanation} & Time complexity for computing the consequent of each antecedent \\ \hline
\multicolumn{1}{c|}{$A$} & \multicolumn{1}{l|}{See detailed explanation} & Number of all feasible antecedents. Usually exponentially increase with $|\mathcal{O}|$ and $L$ (i.e. $|\mathcal{O}|  ^L$) \\ \hline
\multicolumn{1}{c|}{$A'$} & \multicolumn{1}{l|}{See detailed explanation} & Number of sampled antecedents for training of neural consequent estimator \\ \hline

\end{longtable}

\subsection{Extension to Regression Tasks}
\label{sec:regression}
Although we mainly focused on classification tasks, \pname{} can be applied to regression tasks after a small modification. For regression tasks, we change the modeling of neural consequent estimation from a categorical to a direct prediction.
Our neural consequent estimator for regression predicts the value $y'$ instead of $p(y | \bm{\alpha})$ and coverage $c_{\bm{\alpha}}$. Then, we maximize $||y'-y||_2$. 

\subsection{Probability Decomposition}
\label{sec:probability_decomposition}
Here we give the proof for Eq.~(\ref{eq:prob_decomposition_1}): 
\begin{align}
    \label{eq:prob_decomposition_3}
    p(y|\mathbf{x},b) &=
    \sum_{\bm{\alpha}}
    p(y|\bm{\alpha})p(\bm{\alpha}|\mathbf{x},b) \nonumber \\
    &= \sum_{\bm{\alpha}}
    p(y|\bm{\alpha}){p(\bm{\alpha},\mathbf{x},b) \over p(\mathbf{x}, b)} \nonumber \\
    &= \sum_{\bm{\alpha}}
    p(y|\bm{\alpha}) \cdot 
    {p(\bm{\alpha},\mathbf{x},b) \over p(\mathbf{x}, b)} \cdot
    {p(b|\bm{\alpha})p(\bm{\alpha}) \over p(\bm{\alpha}, b)} \cdot
    {p(\bm{\alpha} | \mathbf{x})p(\mathbf{x}) \over p(\bm{\alpha}, \mathbf{x})} \\
  &= \sum_{\bm{\alpha}}
    p(y|\bm{\alpha}) \cdot
    p(b|\bm{\alpha}) \cdot
    p(\bm{\alpha}|\mathbf{x}) \cdot {1 \over p(b)} \cdot
    {{p(\mathbf{x})p(b)} \over {p(\mathbf{x}, b)}} \cdot
    {{p(\bm{\alpha})p(\bm{\alpha}, \mathbf{x}, b)} \over {p(\bm{\alpha}, \mathbf{x})p(\bm{\alpha}, b)}}\\
 &= \sum_{\bm{\alpha}}
    p(y|\bm{\alpha}) \cdot
    p(b|\bm{\alpha}) \cdot
    p(\bm{\alpha}|\mathbf{x}) \cdot {1 \over p(b)} \cdot
    {{p(\mathbf{x})p(b)} \over {p(\mathbf{x}, b)}} \cdot
    {{p(\mathbf{x}, b|\bm{\alpha})} \over {p(\mathbf{x}|\bm{\alpha})p(b|\bm{\alpha})}}\\
    & \propto  \sum_{\bm{\alpha}}
    p(y|\bm{\alpha}) \cdot
    p(b|\bm{\alpha}) \cdot
    p(\bm{\alpha}|\mathbf{x}) \label{eq:proof}
\end{align}

There are two assumptions to hold Eq.~(\ref{eq:proof}).

\textbf{Assumption A}.
For $p(y|\mathbf{x},b) =  \sum_{\bm{\alpha}}p(y|\bm{\alpha})p(\bm{\alpha}|\mathbf{\mathbf{x}},b)$, we assume that $p(y | \bm{\alpha}) = p(y | \bm{\alpha}, \mathbf{x}, b)$.
This is decomposed into two assumptions: $p(y|\bm{\alpha})=p(y|\bm{\alpha},\mathbf{x})$ (A1) and $p(y|\bm{\alpha},\mathbf{x})=p(y|\bm{\alpha},\mathbf{x},b)$ (A2).
\looseness=-1

Assumption A1 $p(y|\bm{\alpha})=p(y|\bm{\alpha},\mathbf{x})$ indicates that explanation $\bm{\alpha}$ contains all information in input $\mathbf{x}$ that is needed to predict $y$. 
This formulation compels the model to pass information from $\mathbf{x}$ to $y$ only via explanations, as opposed to other unexplainable parts. This assumption may limit the prediction performance, but it is essential for $\bm{\alpha}$ to be a trustable explanation for predicting $y$.
Otherwise, there may be a direct connection between $y$ and $\mathbf{x}$ that is unrelated to the explanation $\bm{\alpha}$. Thus, $\bm{\alpha}$ may only explain a small portion of the model behavior (e.g., only explain 1\% of the change in $y$) and differ substantially from the ground-truth explanation of the model behavior. 

Assumption A2 $p(y|\bm{\alpha},\mathbf{x})=p(y|\bm{\alpha},\mathbf{x},b)$ means that  explanation $\bm{\alpha}$ and input $\mathbf{x}$ contain all of the information in $b$ (human prior preference for explanations) that is needed to predict $y$. It is intuitive that this assumption holds, as human preference for explanations is unrelated to the current class label.
\looseness=-1

\textbf{Assumption B}.
For $\sum_{\bm{\alpha}}{p(y|\bm{\alpha})p(\bm{\alpha}|\mathbf{x},b)}\propto\sum_{\bm{\alpha}}{p(b|\bm{\alpha})p(y|\bm{\alpha})p(\bm{\alpha}|\mathbf{x})}$, we assume that $p(\mathbf{x}, b)=p(b)p(\mathbf{x})$ and $p(\mathbf{x}, b|\bm{\alpha})=p(b|\bm{\alpha})p(\mathbf{x}|\bm{\alpha})$.
 
It means that $\mathbf{x}$ and $b$ are independent no matter which $\bm{\alpha}$ is given. In other words, seeing input sample ($\mathbf{x}$) does not change the belief in our prior preference for explanations ($b$), no matter which explanations ($\bm{\alpha}$) are given, i.e., $p(b)=p(b|\mathbf{x})$ and $p(b|\bm{\alpha})=p(b|\mathbf{x},\bm{\alpha})$. The rationale for this assumption is that human preferences for explanations are usually fixed and unrelated with the input $\mathbf{x}$. Even if this assumption is not satisfied, it will not have a significant effect on the framework. Only the human prior module must be integrated into the antecedent generation module, which changes from $p(\bm{\alpha}|\mathbf{x})$ to  $p(\bm{\alpha}|\mathbf{x},b)$.

\subsection{Neural Consequent Estimation}
\label{sec:neural_consequent_estimation}

The input of the neural consequent estimator is the antecedent embedding, which is obtained by $(\mathbf{o}_1...,\mathbf{o}_L)$, where $\mathbf{o}_i$ is the embedding of $o_i$ in $\bm{\alpha}=(o_1...,o_L)$. For each atom, $\mathbf{o}_i$ is initialized as the average embedding of all training samples that satisfy the atom, where the sample embedding can be derived using a pretrained model or $f$. The embeddings of logical connectives are initialized at random, and can be omitted when there is only one logical connective (e.g., AND). We use the Transformer encoder~\cite{vaswani2017attention} as the backbone neural network to emphasize the contextual interaction between atoms and logical connectives.

After encoding $\bm{\alpha}$ with Transformer, an MLP (Multi-Layer Perceptron) layer reduces the representation obtained by mean pooling to a logit. Softmax (multi-class) or sigmoid (two classes) is used to activate the logits to determine the probability for each class $p(y | \bm{\alpha})$ and the coverage $c_{\bm{\alpha}}$ of the antecedent $\bm{\alpha}$, which is converted to the number of observations in the training dataset with $n_{\bm{\alpha}}=c_{\bm{\alpha}}N$. 
The time complexity of deep logic reasoning is significantly reduced by neural estimation of the consequence (Sec.~\ref{sec:optimization}).

The neural consequent estimator is pretrained with $A'=10,000$ sampled rules for each antecedent length (Total $L \times A'$), then used to train the deep antecedent generator with frozen parameters. The following steps are taken to ensure the generality of the rules used in pretraining. To begin, we create the ``true matrix'' (\emph{tm}), that has the size $(|\mathcal{O}| \times N)$, which indicates whether each input sample satisfies each atoms. Then, by multiplying \emph{tm} and its transpose, we can create a matrix of size $(|\mathcal{O}| \times |\mathcal{O}|)$ that indicates the number of samples that satisfy 2-length antecedents ($[o_i, o_j], i, j \in \mathcal{O}$).

Then, we obtain the list of 2-length antecedents whose frequency is larger than a threshold (i.e., \emph{min\_df}). From the 2-length antecedent list, we sample $k \times A'$ rules while $k$ is a hyper-parameter larger than $1$. We set $k$ to be the same with \emph{min\_df} in the experiment. With these $k \times A'$ rules, we can make a new true matrix of size ($(k \times A') \times N$) and repeat the steps to obtain the rules whose frequency is larger than \emph{min\_df}. This sampling process takes linear time to $A'$ instead of $A$, which reduces the time complexity. After the whole process, we can obtain $k \times A'$ number of antecedents for each length. Then we randomly choose $A'$ rules for pretraining of consequent estimator maintaining the balance of labels. In practice, the time spent in sampling process was $1144$s for Yelp, $402$s for Clickbait, and $456$s for Adult dataset in our setting. This time can be even reduced with larger \emph{min\_df}.
\looseness=-1

\subsection{Differentiable Learning}
\label{sec:differentiable_learning}
In Sec.~\ref{sec:optimization}, we sampled one antecedent from $p(\bm{\alpha}^{(s)} | \mathbf{x})$. 
Naive selection (e.g., selecting the maximum value's index) stops the gradient and prevents differential learning of the neural model. 
This problem is solved by sampling $\bm{\alpha}$ with the Straight-Through Gumbel-Softmax function, as shown in Eq.~(\ref{eq:rule_prob}).
For forward propagation, $\bm{\alpha}^{(s)}=(\alpha^{(s)}_1...,\alpha^{(s)}_L)$ is represented by $L$ discrete one-hot vectors. 
To derive $L$ input embeddings for the neural consequent estimator in Sec.~\ref{sec:rule_generator}, each one-hot vector is multiplied by an embedding matrix of atoms and logical connectives.
Differentiable Gumbel-Softmax distribution is used to approximate the gradients during backpropagation.

\subsection{Theoretical Analysis of Explanation Stability}
\label{sec:analysis_explanation_stability}

For linear-regression-based models like SENN~\cite{alvarez2018towards}, the explanations for similar inputs may be entirely different without specific constraints like the robustness loss, because the main optimization goal for SENN is the local prediction accuracy. Without the robustness loss, the model may find a correct prediction locally for a single instance, but being ``surely no more interpretable than any deep neural network'' (quoted from the SENN paper). However, this is not the case for the logic rule reasoning framework, because the antecedent generator is trained to optimize two globally consistent rewards (Eq.~\ref{eq:prob_approximation} and Sec.~\ref{sec:optimization}): human’s prior belief about which explanation types are good and the explanation confidence that is measured by the global prediction accuracy over the entire training dataset given the explanation (logic rule). Thus, explanations for similar inputs may be different only when:
\looseness=-1

\begin{enumerate}
    \item The optimal (most confident and human-preferred) rules for the inputs are different.
    \item There are multiple explanations that achieve the exact same reward.
    \item The model has not been trained sufficiently to achieve the optimal result.
\end{enumerate}

In situation 1), \pname{} removes the heuristic constraint regarding the similarity of explanations, allowing us to identify the optimal explanations for the two inputs. If an instance A is changed to the instance B by substituting “\emph{very disappointing}” with “\emph{disappointing}”, then the best explanation may change from “\emph{very disappointing}” in the instance A to “\emph{awful}” in the instance B. Even if the two instances are similar, their optimal explanations may differ. This is plausible as such a change increases the explanation’s confidence. In other words, the radius of validity of an explanation corresponds to inputs that have similar optimal rules. For example, explanation ``\emph{very disappointing}''$\Rightarrow$\emph{negative sentiment} can generalize to all instances that satisfy the rule and at the same time do not satisfy the more confident rule. When we want to force the explanations of two inputs to be similar, we can also incorporate a constraint that mimics the robustness loss in SENN into the soft human prior. Situation 2) rarely occurs, as our explanation confidence reward  is a real number, not a discrete value. In rare cases where this occurs, it is possible to remedy the situation by using the soft human prior. Situation 3) can be avoided by checking the training loss, the classification accuracy, and the explainability.
\looseness=-1

%% file: source/appendix2_experiment.tex
\section{Supplement for Section 3 (Experiment)}

\subsection{Datasets}
\label{sec:dataset}
We use the following three datasets for experiments. Table~\ref{tab:data_num} reports the number of data points for each dataset that we used for training, validation, and testing.
\textbf{Yelp} classifies reviews of local businesses into positive or negative sentiment~\cite{zhang2015character}. For Yelp, we use a down-sampled subset (10\%) for training, as per existing work~\cite{okajima2019deep}. We split the test dataset and used half of them for the validation dataset.
\textbf{Clickbait News Detection} from Kaggle labels whether a news article is a clickbait~\cite{clickbait2020kaggle}, and we use the ``\emph{news}'' and ``\emph{clickbait}'' classes in the dataset. We split the train data into train and validation.
\textbf{Adult} from the UCI machine learning repository~\cite{uci_ml_repository} is an imbalanced tabular dataset that provides labels about whether the annual income of an adult is more than \$50K/yr or not. We split the data points into train, validation, and test datasets.

\begin{table}[b]
\centering
\caption{
Dataset statistics. The labeling ratio shows whether the data is imbalanced between classes. All data, including training, validation, and test data, is split into the same ratio.
}
\vspace{2mm}
{
\begin{tabular}{lccccc}
\hline
Dataset & \# for training & \# for validation & \# for test & Prediction Labels & Label Ratio \\ \hline
Yelp & 56000 & 19000 & 19000 & Negative, Positive & 1 : 1 \\
Clickbait & 18330 & 1312 & 1312 & News, Clickbait & 3.9 : 1 \\
Adult & 39073 & 4884 & 4885 & \textless{}=50K, \textgreater{}50K & 3.2 : 1 \\ \hline
\end{tabular}
}
\label{tab:data_num}
\vspace{-5mm}
\end{table}

\subsection{Implementation Details}
\label{sec:implementation_details}

\textbf{Hyperparameter settings}.
The backbone models for textual data (i.e., BERT, RoBERTa) follow the original setting, and the model for tabular data (i.e., DNN) consists of network with three fully-connected layers with ReLU activation layers (i.e., FC-ReLU-FC-ReLU-FC) with $512$ hidden dimensions. We employ GRU~\cite{cho2014properties} as a sequential encoder for deep antecedent generation, and Transformer~\cite{vaswani2017attention} as a neural model for consequent estimation, respectively.  For deep antecedent generation, neural consequent estimation, and other baseline models, we set the hidden dimension $|h|$ as the default BERT and RoBERTa embedding size (i.e., $768$) for textual data and $512$ for tabular data.
For training of \pname{}, cross-entropy loss is used for optimization on the probability predicted by the consequent estimator for the antecedents extracted by the antecedent generator.
For RCN, we extract a predefined rule set by following the original work~\cite{okajima2019deep}.
In particular, the predefined rules are decision paths in random forests with $100$ estimators and a maximum depth of four. 
After excluding stopwords, we limit atoms in textual data to only derive from the top-$5000$ most frequent words.
Tabular data uses both categorical and numerical features for atoms while the threshold of numerical features is set to the $25$th, $50$th, $75$th percentiles of data. The length of antecedent $L$ (i.e., the number of atoms from recursive deep antecedent generation) is set to $4$. The minimum document frequency is set to $200$, and the number of rules for pretraining the neural consequent estimator is set to $10,000$.
\looseness=-1

We introduce hyper-parameters in training our model and baselines. Note that the same hyper-parameters are used for training baselines, the neural consequent estimator, and the deep antecedent generator for the all datasets. The base backbone network and self-explainable models are trained 10 epochs. The batch size is set to $16$, the largest size that can be trained on our GPU. For optimization, we employ Adam optimizer with a learning rate of $1e-5$, and ExponentialLR scheduler with $\gamma$ $0.95$. For the learning rate, the one with the best performance is selected after experiments on $5e-5$, $4e-5$, $3e-5$, $2e-5$, and $1e-5$. For SENN, a set of token embeddings from the pretrained language model (i.e., BERT and RoBERTa) are utilized as inputs and are considered to be interpretable basic concepts for textual data experiments. In the case of tabular data, raw input features are used. 
We follow the implementation and hyper-parameter settings for training as in the original work~\cite{alvarez2018towards}. Optimizer or scheduler are also set to be the same as other baselines for a fair comparison.
One NVIDIA A100 is used for each experiment.
\looseness=-1

\textbf{Details about selecting atom candidates}.
We ensure that atoms have a consistent form with baselines for fair comparison.
In current implementation, we only consider atoms that contains the information about existence of a word for given instance (e.g., “\emph{awesome} $\geq$ 1”) for textual datasets. This enables a comparison with explainable models that highlights the words based on their importance weight. We choose top $5000$ frequent words in vocabulary set for atom candidates in main experiments in Sec.~\ref{sec:experiment}. The result with other number of atoms is shown in Sec.~\ref{sec:hyperparameter_sensitivity}. The result with For tabular datasets, we choose different strategies based on feature types. For categorical features, whether the instance belongs to a certain category or not becomes an atom. For example, in the Adult dataset, ``\emph{marital-status} == \emph{Married}''  indicates the person in the given instance is married. For numerical features, we calculate $25$th, $50$th, $75$th percentiles of each feature distribution for the threshold. We use whether the feature of a given sample is larger or smaller than the threshold as atoms to obtain thresholds and use those values to determine the over or under presence of each feature in the given sample. For example, the feature “\emph{age}” of the Adult dataset has thresholds 28, 37,  and 48, which lead to atoms like “\emph{age} $\geq$ 28”, “\emph{age} < 28”, “\emph{age} $\geq$ 37”, “\emph{age} < 37”, “\emph{age} $\geq$ 48”, and “\emph{age} < 48”. This form is consistent with the atoms in our baseline RCN~\cite{okajima2019deep}, which uses random forests for rule creation.
\looseness=-1

It is possible that different atoms associated with the same feature appear in the same explanation,  for example, as in our tabular dataset (e.g., “\emph{age} $\geq$ 37” and “\emph{age} $\geq$ 48” for the feature “\emph{age}”).
In such a situation, we remove the redundant atoms after the explanation has been generated (e.g., removing “\emph{age} $\geq$ 37”). Note that the generated atoms will not be conflicted with each other.
For example, “\emph{age} $\geq$ 48” and “\emph{age} < 37” will not be generated simultaneously in one explanation, because the condition for generation is that the corresponding instance satisfies both atoms.
This is enforced by the local constraint introduced in Sec.~\ref{sec:rule_generator}. We find that such a post-processing step of removing redundant atoms is easy to implement and has reasonably good explainability and prediction performance.
It is also possible to eliminate redundant atoms during explanation generation.
One possible way is to create the atoms so that they do not overlap (e.g., creating “\emph{age} $\geq$ 48”, “48 > \emph{age} $\geq$ 37”, “37 > \emph{age} $\geq$ 28”, “28 > \emph{age}” for feature “\emph{age}”).
However, this may make it impossible to flexibly combine different thresholds (e.g., generating “48 > \emph{age} $\geq$ 28”).
Another way is to apply a mask to the model so that it assigns zero probability to an already chosen feature or a redundant atom.
This can be implemented by carefully setting the local constraint in Sec.~\ref{sec:rule_generator}.

\begin{table}[t]
\centering
\caption{Comparison of classification performance measured in F1. 
The average results from five runs are shown.
The best results among self-explaining models are marked in \textbf{bold}, and the \colorbox{blue!15}{highlighted cells} indicate a similar or better result compared with the unexplainable backbone (Base). The numbers in subscript indicates the standard error of the result.
}

\resizebox{0.9\textwidth}{!}
{
\begin{tabular}{lccccc|c}
\hline
 & \multicolumn{2}{c}{Yelp} & \multicolumn{2}{c}{Clickbait} & Adult &  \\ \cline{2-6}
 & BERT & \multicolumn{1}{c|}{RoBERTa} & BERT & \multicolumn{1}{c|}{RoBERTa} & DNN & \multirow{-2}{*}{Average} \\ \hline
Base & 96.20$_{\ 0.0541}$ & \multicolumn{1}{c|}{97.16$_{\ 0.0672}$} & 72.84$_{\ 0.9302}$ & \multicolumn{1}{c|}{74.25$_{\ 0.7763}$} & 76.15$_{\ 0.2522}$ & 83.32 \\ \hline
SENN & 95.12$_{\ 0.1995}$ & \multicolumn{1}{c|}{96.07$_{\ 0.1180}$} & 69.09$_{\ 0.9550}$ & \multicolumn{1}{c|}{70.99$_{\ 0.5076}$} & 71.69$_{\ 0.7681}$ & 80.59 \\
RCN & \cellcolor[HTML]{CBCEFB}\textbf{96.38$_{\ 0.0089}$} & \multicolumn{1}{c|}{\cellcolor[HTML]{CBCEFB}\textbf{97.36$_{\ 0.0049}$}} & 68.80$_{\ 0.1359}$ & \multicolumn{1}{c|}{68.64$_{\ 0.1467}$} & \cellcolor[HTML]{CBCEFB}\textbf{77.35$_{\ 0.0309}$} & 81.77 \\ \hline
SELOR & \cellcolor[HTML]{CBCEFB}\textbf{96.26$_{\ 0.0445}$} & \multicolumn{1}{c|}{\cellcolor[HTML]{CBCEFB}\textbf{97.13$_{\ 0.0642}$}} & \textbf{71.12$_{\ 0.5479}$} & \multicolumn{1}{c|}{\cellcolor[HTML]{CBCEFB}\textbf{74.20$_{\ 0.5009}$}} & \cellcolor[HTML]{CBCEFB}\textbf{77.37$_{\ 0.0541}$} & \cellcolor[HTML]{CBCEFB}\textbf{83.34} \\ \hline
\end{tabular}
}

\label{tab:prediction_f1}
\end{table}

\subsection{Prediction Performance in F1-score}
We also provide the prediction performance of \pname{} and other self-explainable baselines in Table~\ref{tab:prediction_f1}. The result shows that our method successfully maintains the representation ability of deep learning.

\begin{table}[t]
\centering
\vspace{-5mm}
\caption{
User study results on human precision.
Nine participants P1-P9 were asked to annotate whether an explanation is a reasonable rationale for the prediction. For each compared method, we report the percentage of explanations that are considered good (a, b) or best (c, d).
Avg. and Agr. denote the average and inter-participant agreement, respectively. 
P-values from t-test indicates the statistical significance of the experiment. We mark one star (*) if the p-value is lower than 0.05.
Best results are highlighted in \textbf{bold}.\looseness=-1
}
\vspace{-2mm}

\begin{subtable}[h]{\textwidth}
\centering
\caption{Percentage of good (Yelp)}
\vspace{-2mm}
\resizebox{0.9\textwidth}{!}
{
\begin{tabular}{lccccccccc|c|c|c}
\hline
 & P1 & P2 & P3 & P4 & P5 & P6 & P7 & P8 & P9 & Avg. & Agr. & P-value \\ \hline
\multicolumn{1}{l|}{Lime} & 88 & 82 & \textbf{96} & 90 & 92 & 90 & 98 & 88 & 84 & 89.8 & 84.4 & 8.68 E-04* \\
\multicolumn{1}{l|}{Anchor} & 86 & 74 & 92 & 86 & 84 & 84 & 90 & 78 & 86 & 84.4 & 87.7 & 1.12 E-07* \\
\multicolumn{1}{l|}{SENN} & 26 & 22 & 18 & 32 & 26 & 30 & 80 & 32 & 44 & 34.4 & 72.3 & 1.40 E-51* \\
\multicolumn{1}{l|}{RCN} & 70 & 32 & 6 & 70 & 62 & 74 & 88 & 76 & \textbf{98} & 64.0 & 77.6 & 7.26 E-13* \\ \hline
\multicolumn{1}{l|}{SELOR} & \textbf{90} & \textbf{84} & \textbf{96} & \textbf{98} & \textbf{100} & \textbf{96} & \textbf{100} & \textbf{92} & 94 & \textbf{94.4} & 93.9 & - \\ \hline
\end{tabular}
}
\label{tab:good_yelp_detail}
\end{subtable}

\begin{subtable}[h]{\textwidth}
\centering
\caption{Percentage of good (Adult)}
\vspace{-2mm}
\resizebox{0.9\textwidth}{!}
{
\begin{tabular}{lccccccccc|c|c|c}
\hline
 & P1 & P2 & P3 & P4 & P5 & P6 & P7 & P8 & P9 & Avg. & Agr. & P-value \\ \hline
\multicolumn{1}{l|}{Lime} & \textbf{88} & 24 & 88 & 26 & 76 & 2 & 6 & 26 & 48 & 42.7 & 57.1 & 6.09 E-54* \\
\multicolumn{1}{l|}{Anchor} & 30 & 38 & 32 & 54 & 30 & 84 & 68 & 94 & 44 & 52.7 & 59.9 & 5.56 E-18* \\
\multicolumn{1}{l|}{SENN} & \textbf{88} & 16 & \textbf{90} & 30 & 82 & 4 & 8 & 38 & 58 & 46.0 & 51.5 & 1.18 E-41* \\
\multicolumn{1}{l|}{RCN} & 78 & 70 & 86 & 70 & 56 & 4 & 18 & 86 & 80 & 60.9 & 53.2 & 2.83 E-27* \\ \hline
\multicolumn{1}{l|}{SELOR} & 84 & \textbf{88} & \textbf{90} & \textbf{98} & \textbf{84} & \textbf{98} & \textbf{86} & \textbf{100} & \textbf{88} & \textbf{90.7} & 85.7 & - \\ \hline
\end{tabular}
}
\label{tab:good_adult_detail}
\end{subtable}

\begin{subtable}[h]{\textwidth}
\centering
\vspace{-2mm}
\caption{Percentage of best (Yelp)}
\vspace{-2mm}
\resizebox{0.9\textwidth}{!}
{
\begin{tabular}{lccccccccc|c|c|c}
\hline
 & P1 & P2 & P3 & P4 & P5 & P6 & P7 & P8 & P9 & Avg. & Agr. & P-value \\ \hline
\multicolumn{1}{l|}{Lime} & 30 & 36 & 24 & 40 & 44 & 34 & 14 & \textbf{48} & 38 & 34.2 & 67.6 & 8.87 E-03* \\
\multicolumn{1}{l|}{Anchor} & 24 & 20 & \textbf{36} & 16 & 8 & 12 & 16 & 10 & 20 & 18.0 & 83.6 & 5.63 E-18* \\
\multicolumn{1}{l|}{SENN} & 2 & 4 & 4 & 2 & 2 & 2 & 4 & 0 & 2 & 2.4 & 96.3 & 6.84 E-40* \\
\multicolumn{1}{l|}{RCN} & 2 & 2 & 2 & 0 & 0 & 0 & 6 & 6 & \textbf{0} & 2.0 & 96.3 & 6.84 E-40* \\ \hline
\multicolumn{1}{l|}{SELOR} & \textbf{44} & \textbf{40} & \textbf{36} & \textbf{48} & \textbf{50} & \textbf{54} & \textbf{64} & 40 & \textbf{44} & \textbf{46.7} & 64.8 & - \\ \hline
\end{tabular}
}
\label{tab:best_yelp_detail}
\end{subtable}

\begin{subtable}[h]{\textwidth}
\centering
\vspace{-2mm}
\caption{Percentage of best (Adult)}
\vspace{-2mm}
\resizebox{0.9\textwidth}{!}
{
\begin{tabular}{lccccccccc|c|c|c}
\hline
 & P1 & P2 & P3 & P4 & P5 & P6 & P7 & P8 & P9 & Avg. & Agr. & P-value \\ \hline
\multicolumn{1}{l|}{Lime} & 0 & 2 & 0 & 0 & 8 & 0 & 0 & 0 & 2 & 1.3 & 96.7 & 1.72 E-64* \\
\multicolumn{1}{l|}{Anchor} & 22 & 18 & 20 & 16 & 2 & 2 & 16 & 2 & 26 & 13.8 & 82.9 & 1.23 E-35* \\
\multicolumn{1}{l|}{SENN} & 6 & 10 & 8 & 8 & 34 & 4 & 2 & 2 & 12 & 9.6 & 83.3 & 2.30 E-36* \\
\multicolumn{1}{l|}{RCN} & 10 & 10 & 8 & 12 & 10 & 0 & 6 & 10 & 26 & 10.2 & 82.9 & 1.23 E-35* \\ \hline
\multicolumn{1}{l|}{SELOR} & \textbf{62} & \textbf{60} & \textbf{64} & \textbf{64} & \textbf{46} & \textbf{94} & \textbf{76} & \textbf{86} & \textbf{34} & \textbf{65.1} & 58.4 & - \\ \hline
\end{tabular}
}
\label{tab:best_adult_detail}
\end{subtable}
\vspace{-3mm}
\label{tab:user_study_detail}
\end{table}

\subsection{User Study Details}
\label{sec:user_study_2_detail}

\textbf{Explanation generation process}.
Here, we introduce how we generate the explanations.
\begin{itemize}[nosep,leftmargin=1em,labelwidth=*,align=left]
    \item \textbf{LIME} is distributed as a Python package, and we use \emph{lime\_text} and \emph{lime\_tabular} to generate explanations. The number of disturbances is set to 3,000 for textual data. We choose the words that are consistent with the prediction having positive weights as explanation. To reduce the incongruity with other explanations, we hide the score provided by LIME and join chosen the predicates.

\item \textbf{Anchor} is initialized with an empty set. For every iteration, multiple candidate anchors are produced by extending the current anchor by one additional predicate. Then, the model selects the set of predicates with the highest precision as an anchor while perturbing the other predicates. This process repeats until it satisfies the precision constraint of probability $0.95$.

\item \textbf{SENN} defines the interpretable basis concepts $h(x)$ from the input $\mathbf{x}$, and learns the relevance value $\theta (x)$ which is an interpretable weight in relation to each concept (i.e., $f(x) = \sum_{i}{\theta (x)_i \cdot h(x)_i }$). We choose the set of top-$k$ predicates with the highest positive relevance value as an interpretation for the given input $\mathbf{x}$. $k$ is set to 5. We remove meaningless words (such as ``-'', ``~'') by post-processing. To reduce incongruity with other explanations, we hide the score provided by SENN and join the chosen predicates.

\item \textbf{RCN} chooses a rule from a predefined rule set made by random forest. As the random forest is trained with the bag-of-words of training data, the form of the rule also aligns with the frequency of words.
\looseness=-1

\item \textbf{\pname{}}
recursive deep antecedent generation chooses atoms with the largest weight sequentially. All our atoms are existence of a word (e.g., if ``\emph{good}'' exists), so a rule becomes the list of words. We join these words to explain the given sample.
\end{itemize}

\textbf{User instruction and labeling detail}.
We provided instructions for participants in the form of the guideline file (\emph{Labeling\_Guidelines\_User\_Study\_Table3.pdf}) with detailed description of the task and labeling examples. 
Participants received an Excel file with empty labels, which they were instructed to fill out and return. The snapshot of the Excel file is also attached as a separate file (\emph{Screenshot\_User\_Study\_Table3\_1.PNG}, \emph{Screenshot\_User\_Study\_Table3\_2.PNG}). We originally allowed multiple choices as best explanations, but labelers found it unclear how to decide two explanations are equally good. As this guideline led to confusion and further lower agreement among labelers, we updated the guideline to allow only one best explanation. We conducted the user study twice. During the first survey, we hired three participants. For Yelp dataset, each participant was paid $22.5\$$ per hour with the total budget $45\$$. For Adult dataset, each participant was paid $7.5\$$ per hour with total budget $37.5\$$. At the second survey, we hired six participants, and each participant was paid $7\$$ per hour for both datasets. The total budget we spent in the second survey was $56\$$.

\textbf{Results of all participants}.
Table~\ref{tab:user_study_detail} provides more detailed result including that of each participant.

\textbf{Further discussion on user study results}.
Table~\ref{tab:good_yelp_detail} shows that participants have a low level of agreement on RCN. 
This is because people have varying preferences for the logical connective NOT.
NOT denotes that the prediction is made due to the absence of a particular feature in the text.
One participant (P3) considered most explanations that contained NOT to be noisy because s/he seldomly made decisions based on the absence of a word.

\subsection{Additional Experimental Results}

We describe additional experimental results to support the prediction performance and explanation quality of \pname{}.

\begin{table}[t]
\caption{Performance comparison of \pname{} and a fully transparent model, Random Forest. The backbone model of textual dataset for \pname{} is RoBERTa.}
\label{tab:comparison_transparent}
\centering
\begin{tabular}{lcccccc}
\hline
 & \multicolumn{2}{c}{Yelp} & \multicolumn{2}{c}{Clickbait} & \multicolumn{2}{c}{Adult} \\ \cline{2-7} 
 & F1 & \multicolumn{1}{c|}{AUC} & F1 & \multicolumn{1}{c|}{AUC} & F1 & AUC \\ \hline
Random Forest & 73.03 & \multicolumn{1}{c|}{80.40} & 44.29 & \multicolumn{1}{c|}{60.25} & 65.60 & 66.15 \\
\pname{} & 97.13 & \multicolumn{1}{c|}{97.78} & 74.20 & \multicolumn{1}{c|}{64.14} & 77.37 & 70.36 \\ \hline
\end{tabular}
\end{table}

\begin{figure}[t]
    \centering
    \begin{subfigure}[t]{0.31\textwidth}
        \centering
        \includegraphics[width=\textwidth]{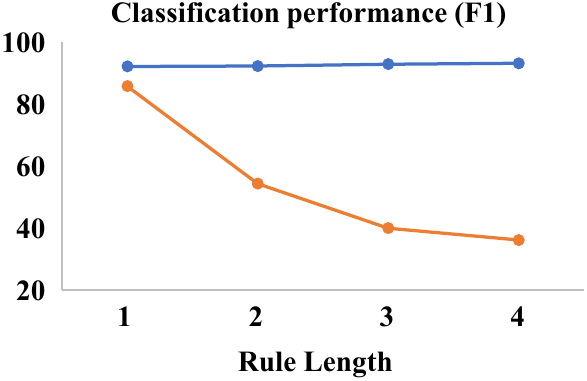}
        \caption{Yelp}
    \end{subfigure}
    \hspace{1mm}
    \begin{subfigure}[t]{0.31\textwidth}
        \centering
        \includegraphics[width=\textwidth]{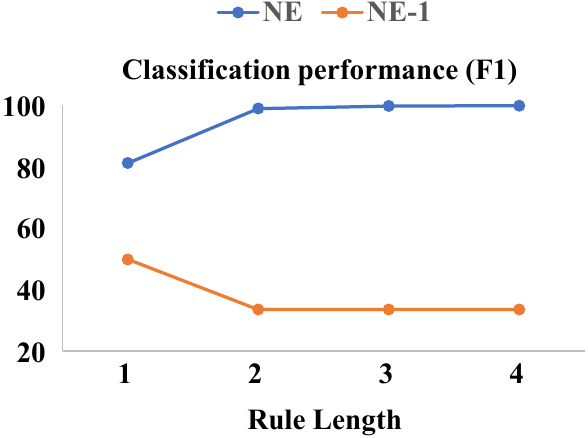}
        \caption{Clickbait}
    \end{subfigure}
    \hspace{1mm}
    \begin{subfigure}[t]{0.31\textwidth}
        \centering
        \includegraphics[width=\textwidth]{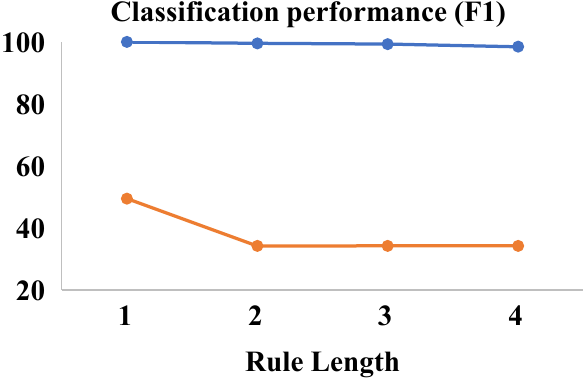}
        \caption{Adult}
    \end{subfigure}
    \caption{The prediction performance of consequent estimator with varying length of antecedents. NE-1 denotes the estimator which is only trained with length-1 antecedents.\looseness=-1}
    \label{fig:consequent_estimator}
\vspace{-3mm}
\end{figure}

\subsubsection{Comparison with Fully Transparent Model}
\label{sec:fully_transparent}

Tree-based models are popular explainable models because their decision process is fully transparent. However, fully transparent models such as decision trees and random forests cannot achieve comparable prediction performance to deep models as shown in Table~\ref{tab:comparison_transparent}.

\begin{wrapfigure}{r}{0.4\textwidth}
    \vspace{-9mm}
	\centering
	\resizebox{0.4\textwidth}{!}
	{
        \includegraphics{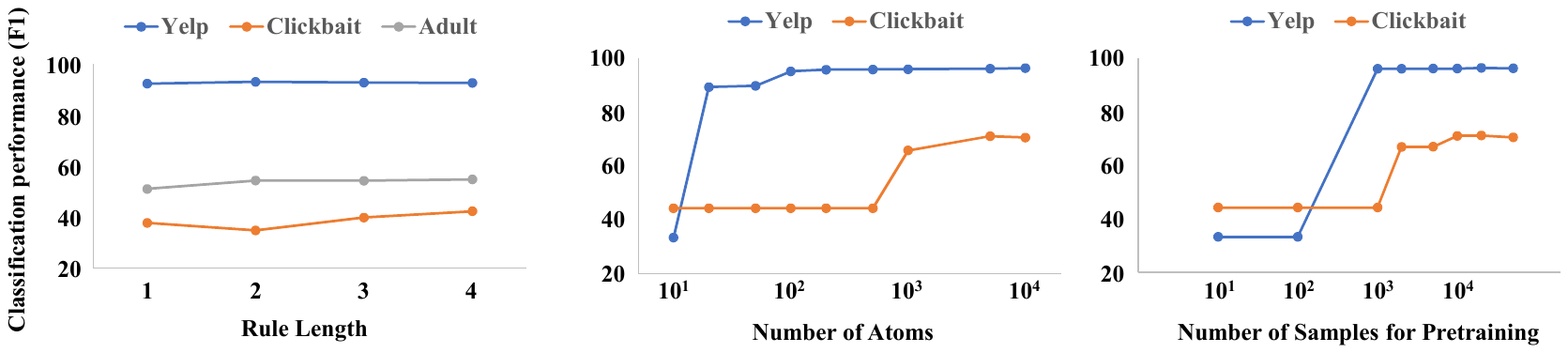}
    }
    \caption{
    The performance of \pname{} with varying number of samples used in pretraining of the neural consequent estimator.
	}
    \vspace{-5mm}
\label{fig:num_samples}
\end{wrapfigure}

\subsubsection{Effectiveness of Neural Consequent Estimator}

\label{sec:effectiveness_consequent_estimator}
The Fig.~\ref{fig:consequent_estimator} shows the prediction performance of our neural consequent estimator (NE) for antecedents of varying length. Our consequent estimator shows reasonable performance in most cases. NE-1 is the estimator that is only pretrained with length-1 antecedents and hence cannot learn the relationship among atoms. Its prediction ability dramatically drops for rules longer than 1.
\looseness=-1

Also, we explore the effect of neural consequent estimator to the overall model performance. Fig.~\ref{fig:num_samples} demonstrates that \pname{} is not highly sensitive to the number of samples used in pretraining the neural consequent estimator, although it requires a minimum level of prediction ability. Additionally, a larger number of samples are needed for more difficult dataset such as Clickbait.
\looseness=-1

\subsubsection{Using Different Logical Connectives}
\label{sec:different_logical_connectives}
We investigate the performance in terms of F1 of different logical connectives on Yelp using BERT as a base model.
First, joining atoms with logical connectives \textbf{OR} leads to a prediction performance of $96.21$, which is similar to the original model using the \textbf{AND} connectives. We also change half of atoms to non-existence rules, which indicates the non-existence of a word (e.g. ``NOT \emph{awesome}'' means the given instance does not contain the word ``\emph{awesome}''). The performance changes to 94.46, and this is natural as the information capacity of non-existence is usually smaller than the existence rules. Additionally, we try \textbf{ORDERED AND}, which considers the order of atoms. For example, ``\emph{not} BEFORE \emph{happy}'' and ``\emph{happy} BEFORE \emph{not}'' will be treated as different antecedents although they have the same words in atoms. Its performance is $96.93$, as the amount of information in the rule increases.

\begin{figure}[b]
    \vspace{5mm}
    \centering
    \begin{subfigure}[t]{0.50\textwidth}
        \centering
        \includegraphics[width=\textwidth]{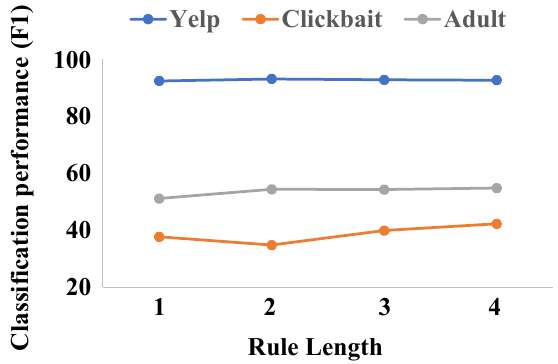}
        \caption{Antecedent Length}
        \label{fig:rule_len}
    \end{subfigure}
    \hspace{1mm}
    \begin{subfigure}[t]{0.46\textwidth}
        \centering
        \includegraphics[width=\textwidth]{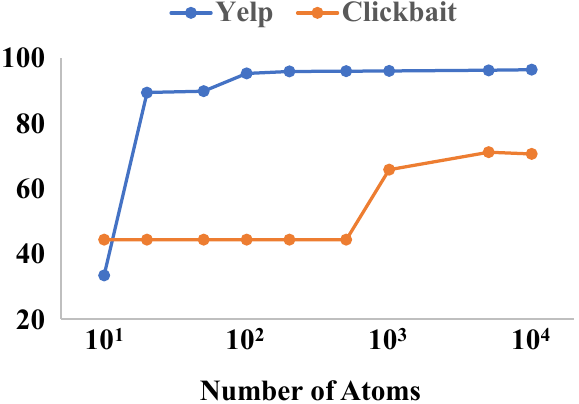}
        \caption{Number of Atoms}
        \label{fig:num_atoms}
    \end{subfigure}

    \caption{The predictive performance of our model with varying antecedent lengths and the number of atoms.\looseness=-1}
    \label{fig:prediction_performance_hyperparameter}
    \vspace{2mm}
\end{figure}

\subsubsection{Hyper-Parameter Sensitivity Analysis}
\label{sec:hyperparameter_sensitivity}

We conduct analysis to test sensitivity of two hyper-parameters: antecedent length $L$ and number of atoms $|\mathcal{O}|$.

\textbf{Impact of hyper-parameters on prediction performance}.
Fig.~\ref{fig:rule_len} shows that \pname{} is not sensitive to the length of antecedent although longer antecedents yield better result in general. Fig~\ref{fig:num_atoms} shows that the number of atoms required for good performance varied by datasets. The more difficult dataset, Clickbait, requires a larger number of atoms to get reasonable performance. However, after certain points, the prediction performance of our method becomes insensitive to number of atoms. 

\textbf{Impact of hyper-parameters on explainability}.
Table~\ref{tab:user_study_length} show how human precision of explanations change with the antecedent length. Antecedents of all lengths, including short antecedents with only one atom, offer a certain level of explainability; The average percentage of good for Length 1 antecedent is 79.7\%. Meanwhile, longer antecedents tend to improve human precision. This indicates the longer antecedents contain more useful information for decision making as it has more chances to find a good atom, resulting in greater precision. Note that the \textit{length of antecedent} is the maximum length of the antecedent; our method can automatically generate shorter antecedents than the default length by electing the NULL atom.
Table~\ref{tab:user_study_num_atoms} shows how human precision of explanations change with the number of candidate atoms. In particular, $1,000$ means that we use the top $1,000$ frequent words as candidate atoms. The explanation quality increases with increasing number of atoms, up to a certain points(i.e., $1,000$ atoms). After this point, there is no statistically significant gain in explainability, demonstrating that \pname{} requires a reasonable size of approximately $1,000$ atoms to provide a good explanation. This finding aligns with the observations in ~\cite{sagar2018how}, which shows that analyzing and explaining text contents such as restaurant reviews and news articles does not require a large vocabulary.

\textbf{Relation between prediction performance and explainability}.
Throughout Fig.~\ref{fig:prediction_performance_hyperparameter}, Table~\ref{tab:user_study_length}, and Table~\ref{tab:user_study_num_atoms}, we could not find concrete evidence for a trade-off between explainability and prediction performance. Rather, we found models with good explainability also produce good prediction performance (i.e., models with antecedent length 2 to 4, models having number of atoms $1,000$ or more atoms). This is consistent with our framework $p(y|\mathbf{x},b)=\sum_{\bm{\alpha}}{p(y|\bm{\alpha})p(\bm{\alpha}|\mathbf{x},b)}$, which passes information from input  to prediction  only via explanations, as opposed to other unexplainable parts. Thus, the expressivity of explanations and the capacity of the model are tightly related. If the hyperparameter settings significantly constrain the expressivity of the explanations (such as limiting the number of atoms to 10), both explanation quality and predictive performance will decrease significantly.

\begin{table}[t]
\centering
\vspace{-5mm}
\caption{
User study results on human precision with varying antecedent lengths.
Participants P1-P6 were asked to annotate whether an explanation is a reasonable rationale for the prediction. For each length, we report the percentage of explanations that are considered good (a) or best (b).
Avg. and Agr. denote the average and inter-participant agreement, respectively. 
P-values from t-test indicates the statistical significance of the experiment. We mark one star (*) if the p-value is lower than 0.05.
Best results are highlighted in \textbf{bold}.\looseness=-1
}
\vspace{-2mm}

\begin{subtable}[h]{\textwidth}
\centering
\caption{Percentage of good}
\vspace{-2mm}
\resizebox{0.75\textwidth}{!}
{
\begin{tabular}{l|cccccc|c|c|c}
\hline
 & P1 & P2 & P3 & P4 & P5 & P6 & Avg. & Agr. & P-value \\ \hline
Length 1 & 76 & 74 & 76 & 76 & 96 & 80 & 79.7 & 87.6 & 1.31 E-11* \\
Length 2 & 92 & 90 & 86 & 94 & \textbf{100} & 84 & 91.0 & 91.3 & 8.93 E-04* \\
Length 3 & 96 & 94 & 88 & 92 & \textbf{100} & 84 & 92.3 & 89.7 & 3.73 E-03* \\
Length 4 & \textbf{100} & \textbf{100} & \textbf{90} & \textbf{98} & \textbf{100} & \textbf{86} & \textbf{95.6} & 91.9 & - \\ \hline
\end{tabular}
}
\label{tab:good_length}
\end{subtable}

\begin{subtable}[h]{\textwidth}
\centering
\caption{Percentage of best}
\vspace{-2mm}
\resizebox{0.75\textwidth}{!}
{
\begin{tabular}{l|cccccc|c|c|c}
\hline
 & P1 & P2 & P3 & P4 & P5 & P6 & Avg. & Agr. & P-value \\ \hline
Length 1 & 8 & 4 & 2 & 6 & 10 & 8 & 6.3 & 94.3 & 2.38 E-43* \\
Length 2 & 24 & 26 & 10 & 12 & 14 & 10 & 16.0 & 81.6 & 6.98 E-23* \\
Length 3 & 20 & 28 & 18 & 28 & 14 & 26 & 22.3 & 81.7 & 1.58 E-16* \\
Length 4 & \textbf{56} & \textbf{50} & \textbf{68} & \textbf{62} & \textbf{72} & \textbf{58} & \textbf{61.0} & 66.5 & - \\ \hline
\end{tabular}
}
\label{tab:best_length}
\end{subtable}
\vspace{-3mm}
\label{tab:user_study_length}
\end{table}

\begin{table}[t]
\centering
\caption{
User study results on human precision with varying number of atoms.
Participants P1-P6 were asked to annotate whether an explanation is a reasonable rationale for the prediction. For each length, we report the percentage of explanations that are considered good (a) or best (b).
Avg. and Agr. denote the average and inter-participant agreement, respectively. 
P-values from t-test indicate the statistical significance of the experiment. We mark one star (*) if the p-value is lower or close to 0.05.
Best results are highlighted in \textbf{bold}.\looseness=-1
}
\vspace{-2mm}

\begin{subtable}[h]{\textwidth}
\centering
\caption{Percentage of good}
\vspace{-2mm}
\resizebox{0.75\textwidth}{!}
{
\begin{tabular}{l|cccccc|c|c|c}
\hline
\# Atoms & P1 & P2 & P3 & P4 & P5 & P6 & Avg. & Agr. & P-value \\ \hline
10 & 10 & 12 & 14 & 8 & 26 & 24 & 15.7 & 94.5 & 5.86 E-02* \\
100 & 34 & 40 & 46 & 34 & 64 & 58 & 46.0 & 94.5 & 5.86 E-02* \\
1000 & 84 & 82 & 86 & 78 & \textbf{100} & 80 & 85.0 & 94.1 & 1.58 E-01 \\
5000 & \textbf{100} & \textbf{98} & \textbf{98} & \textbf{90} & \textbf{100} & 86 & \textbf{95.3} & 91.7 & \textbf{-} \\
10000 & 96 & 96 & 94 & 88 & \textbf{100} & \textbf{90} & 94.0 & 91.7 & 2.06 E-01 \\ \hline
\end{tabular}
}
\label{tab:good_num_atoms}
\end{subtable}

\begin{subtable}[h]{\textwidth}
\centering
\caption{Percentage of best}
\vspace{-2mm}
\resizebox{0.75\textwidth}{!}
{
\begin{tabular}{l|cccccc|c|c|c}
\hline
\# Atoms & P1 & P2 & P3 & P4 & P5 & P6 & Avg. & Agr. & P-value \\ \hline
10 & 0 & 2 & 4 & 0 & 0 & 0 & 1.0 & 77.1 & 6.25 E-5** \\
100 & 6 & 6 & 4 & 6 & 4 & 2 & 4.7 & 76.9 & 8.16 E-5** \\
1000 & 30 & 38 & 32 & 30 & 44 & 34 & 34.7 & 73.5 & 2.45 E-3** \\
5000 & \textbf{54} & \textbf{56} & \textbf{56} & \textbf{52} & \textbf{48} & \textbf{50} & \textbf{52.7} & 71.2 & \textbf{-} \\
10000 & 46 & 48 & 52 & 46 & \textbf{48} & 46 & 47.7 & 74.3 & 2.20 E-1 \\ \hline
\end{tabular}
}
\label{tab:best_num_atoms}
\end{subtable}
\vspace{-3mm}
\label{tab:user_study_num_atoms}
\end{table}


\newpage
\subsection{Explanation Stability}
\label{sec:explanation_stability}

\textbf{Do explanations keep the same in different runs?}
We conduct experiments to confirm that our model usually generates unique explanations for the same instances in different runs. Comparing the model explanations trained with 5 seeds reveals that, on average, 90.04\% of atoms were shared by explanations from different seeds, and 71.27\% were identical on Yelp. This comparison suggests that our model generates a unique explanation for the same instance, even in the absence of a direct controlling factor. The reason why we can generate unique explanations is that we optimize the explanation generator with two globally consistent rewards in Eq.~\ref{eq:prob_decomposition_2}: 1) human’s prior belief about which explanation types are good and 2) the explanation (rule) confidence that is measured by the global prediction accuracy over the entire training corpus given the rule. Since the second reward is a real number instead of a discrete value and has a globally consistent meaning, the optimal explanation is usually unique and stable, leading to similar results when trained with different random seeds.

\textbf{Do similar instances lead to similar explanations?}
Table~\ref{tab:explanation_stability} shows examples of generated explanations for similar inputs. \pname{} successfully maintains its explanation when  minor changes are made to input words, but suggests a new explanation when critical changes are made. In case (a), for example, our method provides the same explanation when the words ``\emph{pizza}'' and ``\emph{waiters}'' are changed to ``\emph{pasta}'' and ``\emph{servers}''. However, when sentiment-related words such as ``\emph{cold}'' and ``\emph{rude}'' are changed, it adapts to the new words and gives a new explanation.

\begin{longtable}{cccc}
\caption{Generated explanations of samples and their perturbation. The manually changed words are highlighted in \textbf{bold}}
\label{tab:explanation_stability} \\

\hline
Case & \multicolumn{1}{c}{Sample} & Model Explanation & Prediction \\ \hline
\multicolumn{1}{c|}{\multirow{3}{*}{}} & \multicolumn{1}{m{0.5\textwidth}|}{This place is awful. The pizza was cold, and the waiters were rude.} & \multicolumn{1}{>{\centering\arraybackslash}m{0.2\textwidth}|}{awful, cold, rude} & Negative \\ \cline{2-4} 
\multicolumn{1}{c|}{(a)} & \multicolumn{1}{m{0.5\textwidth}|}{This place is awful. The \textbf{pasta} was cold, and the \textbf{servers} were rude.} & \multicolumn{1}{>{\centering\arraybackslash}m{0.2\textwidth}|}{awful, cold, rude} & Negative \\ \cline{2-4} 
\multicolumn{1}{c|}{} & \multicolumn{1}{m{0.5\textwidth}|}{This place is awful. The pizza was \textbf{undercooked}, and the waiters were \textbf{unfriendly}.} & \multicolumn{1}{>{\centering\arraybackslash}m{0.2\textwidth}|}{awful, undercooked, unfriendly} & Negative \\ \hline
\multicolumn{1}{c|}{\multirow{3}{*}{}} & \multicolumn{1}{m{0.5\textwidth}|}{I love here. It was an amazing experience to eat a cheesy macaroni.} & \multicolumn{1}{>{\centering\arraybackslash}m{0.2\textwidth}|}{love, amazing, cheesy} & Positive \\ \cline{2-4} 
\multicolumn{1}{c|}{(b)} & \multicolumn{1}{m{0.5\textwidth}|}{I \textbf{recommend} here. It was \textbf{a happy} experience to eat a cheesy macaroni.} & \multicolumn{1}{>{\centering\arraybackslash}m{0.2\textwidth}|}{recommend, happy, cheesy} & Positive \\ \cline{2-4} 
\multicolumn{1}{c|}{} & \multicolumn{1}{m{0.5\textwidth}|}{I \textbf{hate} here. It was \textbf{a bad} experience to eat a cheesy macaroni.} & \multicolumn{1}{>{\centering\arraybackslash}m{0.2\textwidth}|}{hate, bad, experience} & Negative \\ \hline
\multicolumn{1}{c|}{\multirow{3}{*}{}} & \multicolumn{1}{m{0.5\textwidth}|}{I ordered three tacos and all 3 were downright lousy. Can't remember the last time I had food this bad. The shrimp taco was overbreaded and in a sickly sweet sauce, the shredded beef taco was very tiny and thankfully, I can't remember what the third taco tasted like. To the reviewer who posted that these tacos are top notch.... what are you smoking? I waited forever to get my food and saw numerous other people who came in after me get their food. Waiter was MIA. Not coming back....ever.} & \multicolumn{1}{>{\centering\arraybackslash}m{0.2\textwidth}|}{lousy, bad, waited, forever} & Negative \\ \cline{2-4} 
\multicolumn{1}{c|}{(c)} & \multicolumn{1}{m{0.5\textwidth}|}{I ordered three tacos and all 3 were downright lousy. Can't remember the last time I had food this bad. The shrimp taco was overbreaded and in a sickly sweet sauce, the shredded beef taco was very tiny and thankfully, I can't remember what the third taco tasted like. To the reviewer who posted that these tacos are top notch.... what are you smoking? I waited \textbf{a little} to get my food and saw numerous other people who came in after me get their food. Waiter was MIA. Not coming back....ever.} & \multicolumn{1}{>{\centering\arraybackslash}m{0.2\textwidth}|}{lousy, bad, waited, not} & Negative \\ \cline{2-4} 
\multicolumn{1}{c|}{} & \multicolumn{1}{m{0.5\textwidth}|}{I ordered three \textbf{awful, terrible} tacos and all 3 were downright lousy. Can't remember the last time I had food this bad. The shrimp taco was overbreaded and in a sickly sweet sauce, the shredded beef taco was very tiny and thankfully, I can't remember what the third taco tasted like. To the reviewer who posted that these tacos are top notch.... what are you smoking? I waited forever to get my food and saw numerous other people who came in after me get their food. Waiter was MIA. Not coming back....ever.} & \multicolumn{1}{>{\centering\arraybackslash}m{0.2\textwidth}|}{awful, terrible, waited, forever} & Negative \\ \hline
\multicolumn{1}{c|}{\multirow{3}{*}{}} & \multicolumn{1}{m{0.5\textwidth}|}{I had an amazing 4 course meal here with my family from philadlephia. my father runs a farmers market there and was very impressed with their use of seasonal and local foods. We had an amazing pork belly salad and I had duck wrapped in bacon and stuffed with pate, which sounds insanely heavy, but it was not; the portion was small enough not to be overwhelmed and it was not overly greasy at all. It was a fantastic meal. I think l'etoile is on par with top restaurants in bigger cities.} & \multicolumn{1}{>{\centering\arraybackslash}m{0.2\textwidth}|}{amazing, family, stuffed, fantastic} & Positive \\ \cline{2-4} 
\multicolumn{1}{c|}{(d)} & \multicolumn{1}{m{0.5\textwidth}|}{I had \textbf{a great} 4 course meal here with my family from philadlephia. my father runs a farmers market there and was very impressed with their use of seasonal and local foods. We had an amazing pork belly salad and I had duck wrapped in bacon and stuffed with pate, which sounds insanely heavy, but it was not; the portion was small enough not to be overwhelmed and it was not overly greasy at all. It was a fantastic meal. I think l'etoile is on par with top restaurants in bigger cities.} & \multicolumn{1}{>{\centering\arraybackslash}m{0.2\textwidth}|}{great, family, amazing, fantastic} & Positive \\ \cline{2-4} 
\multicolumn{1}{c|}{} & \multicolumn{1}{m{0.5\textwidth}|}{I had an \textbf{awful} 4 course meal here with my family from philadlephia. my father runs a farmers market there and was very \textbf{disappointed} with their use of seasonal and local foods. We had a \textbf{terrible} pork belly salad and I had duck wrapped in bacon and stuffed with pate, which sounds insanely heavy; \textbf{and it was right}; the portion was \textbf{too small to be full} and it \textbf{was overly} greasy at all. It was a \textbf{bad} meal. I think l'etoile is on par with \textbf{bad} restaurants in bigger cities.} & \multicolumn{1}{>{\centering\arraybackslash}m{0.2\textwidth}|}{awful, disappointed, terrible, bad} & Negative \\ \hline

\end{longtable}

%% file: main.bbl
\begin{thebibliography}{10}

\bibitem{ribeiro2016should}
Marco~Tulio Ribeiro, Sameer Singh, and Carlos Guestrin.
\newblock {"Why Should I Trust You?"} explaining the predictions of any
  classifier.
\newblock In {\em KDD}, 2016.

\bibitem{thrun1994extracting}
Sebastian Thrun.
\newblock Extracting rules from artificial neural networks with distributed
  representations.
\newblock In {\em NeurIPS}, 1994.

\bibitem{ribeiro2016nothing}
Marco~Tulio Ribeiro, Sameer Singh, and Carlos Guestrin.
\newblock Nothing else matters: Model-agnostic explanations by identifying
  prediction invariance.
\newblock {\em stat}, 2016.

\bibitem{lei2016rationalizing}
Tao Lei, Regina Barzilay, and Tommi Jaakkola.
\newblock Rationalizing neural predictions.
\newblock In {\em EMNLP}, 2016.

\bibitem{alikaniotis2016automatic}
Dimitrios Alikaniotis, Helen Yannakoudakis, and Marek Rei.
\newblock Automatic text scoring using neural networks.
\newblock In {\em ACL}, 2016.

\bibitem{strobelt2017lstmvis}
Hendrik Strobelt, Sebastian Gehrmann, Hanspeter Pfister, and Alexander~M Rush.
\newblock Lstmvis: A tool for visual analysis of hidden state dynamics in
  recurrent neural networks.
\newblock {\em IEEE TVCG}, 2017.

\bibitem{murdoch2018beyond}
W~James Murdoch, Peter~J Liu, and Bin Yu.
\newblock Beyond word importance: Contextual decomposition to extract
  interactions from lstms.
\newblock In {\em ICLR}, 2018.

\bibitem{peake2018explanation}
Georgina Peake and Jun Wang.
\newblock Explanation mining: Post hoc interpretability of latent factor models
  for recommendation systems.
\newblock In {\em KDD}, 2018.

\bibitem{liang2020adversarial}
Jian Liang, Bing Bai, Yuren Cao, Kun Bai, and Fei Wang.
\newblock Adversarial infidelity learning for model interpretation.
\newblock In {\em KDD}, 2020.

\bibitem{gao2021learning}
Jingyue Gao, Xiting Wang, Yasha Wang, Yulan Yan, and Xing Xie.
\newblock Learning groupwise explanations for black-box models.
\newblock In {\em IJCAI}, 2021.

\bibitem{alvarez2018towards}
David Alvarez~Melis and Tommi Jaakkola.
\newblock Towards robust interpretability with self-explaining neural networks.
\newblock In {\em NeurIPS}, 2018.

\bibitem{rudin2019stop}
Cynthia Rudin.
\newblock Stop explaining black box machine learning models for high stakes
  decisions and use interpretable models instead.
\newblock {\em Nature Machine Intelligence}, 2019.

\bibitem{lundberg2017unified}
Scott~M Lundberg and Su-In Lee.
\newblock A unified approach to interpreting model predictions.
\newblock In {\em NeurIPS}, 2017.

\bibitem{ribeiro2018anchors}
Marco~Tulio Ribeiro, Sameer Singh, and Carlos Guestrin.
\newblock Anchors: High-precision model-agnostic explanations.
\newblock In {\em AAAI}, 2018.

\bibitem{guan2019towards}
Chaoyu Guan, Xiting Wang, Quanshi Zhang, Runjin Chen, Di~He, and Xing Xie.
\newblock Towards a deep and unified understanding of deep neural models in
  nlp.
\newblock In {\em ICML}, 2019.

\bibitem{hong2020human}
Sungsoo~Ray Hong, Jessica Hullman, and Enrico Bertini.
\newblock Human factors in model interpretability: Industry practices,
  challenges, and needs.
\newblock {\em PACM HCI}, 2020.

\bibitem{letham2015interpretable}
Benjamin Letham, Cynthia Rudin, Tyler~H McCormick, and David Madigan.
\newblock Interpretable classifiers using rules and bayesian analysis: Building
  a better stroke prediction model.
\newblock {\em AOAS}, 2015.

\bibitem{yang2017scalable}
Hongyu Yang, Cynthia Rudin, and Margo Seltzer.
\newblock Scalable bayesian rule lists.
\newblock In {\em ICML}, 2017.

\bibitem{evans2018learning}
Richard Evans and Edward Grefenstette.
\newblock Learning explanatory rules from noisy data.
\newblock {\em Journal of Artificial Intelligence Research}, 2018.

\bibitem{angelino2017learning}
Elaine Angelino, Nicholas Larus-Stone, Daniel Alabi, Margo Seltzer, and Cynthia
  Rudin.
\newblock Learning certifiably optimal rule lists.
\newblock In {\em KDD}, 2017.

\bibitem{ming2019interpretable}
Yao Ming, Panpan Xu, Huamin Qu, and Liu Ren.
\newblock Interpretable and steerable sequence learning via prototypes.
\newblock In {\em KDD}, 2019.

\bibitem{chen2020towards}
Zhongxia Chen, Xiting Wang, Xing Xie, Mehul Parsana, Akshay Soni, Xiang Ao, and
  Enhong Chen.
\newblock Towards explainable conversational recommendation.
\newblock In {\em IJCAI}, 2020.

\bibitem{kulesza2015principles}
Todd Kulesza, Margaret Burnett, Weng-Keen Wong, and Simone Stumpf.
\newblock Principles of explanatory debugging to personalize interactive
  machine learning.
\newblock In {\em IUI}, 2015.

\bibitem{schramowski2020making}
Patrick Schramowski, Wolfgang Stammer, Stefano Teso, Anna Brugger, Franziska
  Herbert, Xiaoting Shao, Hans-Georg Luigs, Anne-Katrin Mahlein, and Kristian
  Kersting.
\newblock Making deep neural networks right for the right scientific reasons by
  interacting with their explanations.
\newblock {\em Nature Machine Intelligence}, 2020.

\bibitem{lertvittayakumjorn2020find}
Piyawat Lertvittayakumjorn, Lucia Specia, and Francesca Toni.
\newblock Find: Human-in-the-loop debugging deep text classifiers.
\newblock In {\em EMNLP}, 2020.

\bibitem{ciravegna2021human}
Gabriele Ciravegna, Francesco Giannini, Marco Gori, Marco Maggini, and Stefano
  Melacci.
\newblock Human-driven fol explanations of deep learning.
\newblock In {\em IJCAI}, 2021.

\bibitem{stammer2021right}
Wolfgang Stammer, Patrick Schramowski, and Kristian Kersting.
\newblock Right for the right concept: Revising neuro-symbolic concepts by
  interacting with their explanations.
\newblock In {\em CVPR}, 2021.

\bibitem{bontempelli2021toward}
Andrea Bontempelli, Fausto Giunchiglia, Andrea Passerini, and Stefano Teso.
\newblock Toward a unified framework for debugging gray-box models.
\newblock {\em arXiv preprint arXiv:2109.11160}, 2021.

\bibitem{de2020statistical}
Luc De~Raedt, Sebastijan Dumancic, Robin Manhaeve, and Giuseppe Marra.
\newblock From statistical relational to neuro-symbolic artificial
  intelligence.
\newblock In {\em IJCAI}, 2020.

\bibitem{valkov2018houdini}
Lazar Valkov, Dipak Chaudhari, Akash Srivastava, Charles Sutton, and Swarat
  Chaudhuri.
\newblock Houdini: lifelong learning as program synthesis.
\newblock In {\em NeurIPS}, 2018.

\bibitem{ellis2018learninglibraries}
Kevin Ellis, Lucas Morales, Mathias Sabl{\'e}-Meyer, Armando Solar-Lezama, and
  Josh Tenenbaum.
\newblock Learning libraries of subroutines for neurally-guided bayesian
  program induction.
\newblock In {\em NeurIPS}, 2018.

\bibitem{kalyan2018neural}
Ashwin Kalyan, Abhishek Mohta, Oleksandr Polozov, Dhruv Batra, Prateek Jain,
  and Sumit Gulwani.
\newblock Neural-guided deductive search for real-time program synthesis from
  examples.
\newblock In {\em ICLR}, 2018.

\bibitem{ellis2018learningto}
Kevin Ellis, Daniel Ritchie, Armando Solar-Lezama, and Josh Tenenbaum.
\newblock Learning to infer graphics programs from hand-drawn images.
\newblock In {\em NeurIPS}, 2018.

\bibitem{mao2019neuro}
Jiayuan Mao, Chuang Gan, Pushmeet Kohli, Joshua~B Tenenbaum, and Jiajun Wu.
\newblock The neuro-symbolic concept learner: Interpreting scenes, words, and
  sentences from natural supervision.
\newblock In {\em ICLR}, 2019.

\bibitem{yu2022probabilistic}
Dongran Yu, Bo~Yang, Qianhao Wei, Anchen Li, and Shirui Pan.
\newblock A probabilistic graphical model based on neural-symbolic reasoning
  for visual relationship detection.
\newblock In {\em CVPR}, 2022.

\bibitem{wang2022multi}
Xiting Wang, Kunpeng Liu, Dongjie Wang, Le~Wu, Yanjie Fu, and Xing Xie.
\newblock Multi-level recommendation reasoning over knowledge graphs with
  reinforcement learning.
\newblock In {\em WebConf}, 2022.

\bibitem{yang2017differentiable}
Fan Yang, Zhilin Yang, and William~W Cohen.
\newblock Differentiable learning of logical rules for knowledge base
  reasoning.
\newblock In {\em NeurIPS}, 2017.

\bibitem{zhao2020leveraging}
Kangzhi Zhao, Xiting Wang, Yuren Zhang, Li~Zhao, Zheng Liu, Chunxiao Xing, and
  Xing Xie.
\newblock Leveraging demonstrations for reinforcement recommendation reasoning
  over knowledge graphs.
\newblock In {\em SIGIR}, 2020.

\bibitem{okajima2019deep}
Yuzuru Okajima and Kunihiko Sadamasa.
\newblock Deep neural networks constrained by decision rules.
\newblock In {\em AAAI}, 2019.

\bibitem{jang2017categorical}
Eric Jang, Shixiang Gu, and Ben Poole.
\newblock Categorical reparameterization with gumbel-softmax.
\newblock {\em stat}, 2017.

\bibitem{kendall2018multi}
Alex Kendall, Yarin Gal, and Roberto Cipolla.
\newblock Multi-task learning using uncertainty to weigh losses for scene
  geometry and semantics.
\newblock In {\em CVPR}, 2018.

\bibitem{cho2014properties}
Kyunghyun Cho, Bart van Merri{\"e}nboer, Dzmitry Bahdanau, and Yoshua Bengio.
\newblock On the properties of neural machine translation: Encoder--decoder
  approaches.
\newblock In {\em SSST}, 2014.

\bibitem{vaswani2017attention}
Ashish Vaswani, Noam Shazeer, Niki Parmar, Jakob Uszkoreit, Llion Jones,
  Aidan~N Gomez, {\L}ukasz Kaiser, and Illia Polosukhin.
\newblock Attention is all you need.
\newblock In {\em NeurIPS}, 2017.

\bibitem{zhang2015character}
Xiang Zhang, Junbo Zhao, and Yann LeCun.
\newblock Character-level convolutional networks for text classification.
\newblock In {\em NeurIPS}, 2015.

\bibitem{clickbait2020kaggle}
Open Data~Science (\url{ODS.ai}).
\newblock {Kaggle} clickbait news detection.
\newblock \url{https://www.kaggle.com/c/clickbait-news-detection}, 2020.

\bibitem{uci_ml_repository}
Dheeru Dua and Casey Graf.
\newblock {UCI} machine learning repository.
\newblock \url{http://archive.ics.uci.edu/ml}, 2017.

\bibitem{saito2015precision}
Takaya Saito and Marc Rehmsmeier.
\newblock The precision-recall plot is more informative than the roc plot when
  evaluating binary classifiers on imbalanced datasets.
\newblock {\em PloS one}, 10(3):e0118432, 2015.

\bibitem{kenton2019bert}
Jacob Devlin Ming-Wei~Chang Kenton and Lee~Kristina Toutanova.
\newblock Bert: Pre-training of deep bidirectional transformers for language
  understanding.
\newblock In {\em NAACL}, 2019.

\bibitem{liu2019roberta}
Yinhan Liu, Myle Ott, Naman Goyal, Jingfei Du, Mandar Joshi, Danqi Chen, Omer
  Levy, Mike Lewis, Luke Zettlemoyer, and Veselin Stoyanov.
\newblock Roberta: A robustly optimized bert pretraining approach.
\newblock {\em arXiv preprint arXiv:1907.11692}, 2019.

\bibitem{li2019dividemix}
Junnan Li, Richard Socher, and Steven~CH Hoi.
\newblock Dividemix: Learning with noisy labels as semi-supervised learning.
\newblock In {\em ICLR}, 2019.

\bibitem{zhang2018mixup}
Hongyi Zhang, Moustapha Cisse, Yann~N Dauphin, and David Lopez-Paz.
\newblock Mixup: Beyond empirical risk minimization.
\newblock In {\em ICLR}, 2018.

\bibitem{speechocean}
Speechocean.
\newblock \url{https://en.speechocean.com/}.

\bibitem{kim2018interpretability}
Been Kim, Martin Wattenberg, Justin Gilmer, Carrie Cai, James Wexler, Fernanda
  Viegas, et~al.
\newblock Interpretability beyond feature attribution: Quantitative testing
  with concept activation vectors (tcav).
\newblock In {\em ICML}, 2018.

\bibitem{peer2022data}
Eyal Peer, David Rothschild, Andrew Gordon, Zak Evernden, and Ekaterina Damer.
\newblock Data quality of platforms and panels for online behavioral research.
\newblock {\em Behavior Research Methods}, 54(4):1643--1662, 2022.

\bibitem{sagar2018how}
Beth Sagar-Fenton and Lizzy McNeill.
\newblock How many words do you need to speak a language.
\newblock {\em BBC}, 2018.

\end{thebibliography}
